\documentclass[conference]{IEEEtran}
\usepackage{enumitem}
\IEEEoverridecommandlockouts
\usepackage{cite}
\usepackage{amsmath,amssymb,amsfonts}
\usepackage{algorithmic}
\usepackage{graphicx}
\usepackage{textcomp}
\usepackage{xcolor}
\usepackage{amsmath}
\usepackage{amssymb}
\usepackage{mathtools}
\usepackage{amsthm}
\usepackage{microtype}
\usepackage{graphicx}
\usepackage{booktabs} 
\usepackage{enumitem}
\usepackage{float}
\usepackage{multirow}
\usepackage{subcaption}

\newcommand{\Reg}[0]{{\rm Reg}}

\usepackage{hyperref}
\usepackage{enumitem}
\usepackage[boxed, ruled, vlined, linesnumbered]{algorithm2e}

\def\BibTeX{{\rm B\kern-.05em{\sc i\kern-.025em b}\kern-.08em
    T\kern-.1667em\lower.7ex\hbox{E}\kern-.125emX}}
\begin{document}

\title{MGAug: Multimodal Geometric Augmentation in \\ Latent Spaces of Image Deformations\\
}

\author{\IEEEauthorblockN{Tonmoy Hossain and Miaomiao Zhang}
\IEEEauthorblockA{\textit{Department of CS and ECE} \\ 
\textit{University of Virginia}\\
Charlottesville, VA, USA}
}

\maketitle
\pagestyle{plain}

\begin{abstract}
Geometric transformations have been widely used to augment the size of training images. Existing methods often assume a unimodal distribution of the underlying transformations between images, which limits their power when data with multimodal distributions occur. In this paper, we propose a novel model, {\em Multimodal Geometric Augmentation} (MGAug), that for the first time generates augmenting transformations in a multimodal latent space of geometric deformations. To achieve this, we first develop a deep network that embeds the learning of latent geometric spaces of diffeomorphic transformations (a.k.a. diffeomorphisms) in a variational autoencoder (VAE). A mixture of multivariate Gaussians is formulated in the tangent space of diffeomorphisms and serves as a prior to approximate the hidden distribution of image transformations. We then augment the original training dataset by deforming images using randomly sampled transformations from the learned multimodal latent space of VAE. To validate the efficiency of our model, we jointly learn the augmentation strategy with two distinct domain-specific tasks: multi-class classification on 2D synthetic datasets and segmentation on real 3D brain magnetic resonance images (MRIs). We also compare MGAug with state-of-the-art transformation-based image augmentation algorithms. Experimental results show that our proposed approach outperforms all baselines by significantly improved prediction accuracy. Our code is publicly available at \href{https://github.com/tonmoy-hossain/MGAug}{GitHub}.
\end{abstract}

\section{Introduction}
\label{sec:intro}
Recent advances in deep learning (DL) have led to remarkable progress in image analysis tasks, including but not limited to image classification \cite{krizhevsky2017imagenet,wang2022geo,hossain2024invariant}, segmentation~\cite{hossain2019brain,chen2021transunet}, object detection~\cite{kim2016shape,pang2019libra}, and 3D reconstruction~\cite{jayakumar2023sadir}. The literature has shown that deep neural networks (DNNs) are most powerful when trained on vast amounts of manually labeled data. Unfortunately, many image applications have relatively small annotated datasets available, due to privacy concerns, regulations, costs, and incidents of disease. The DNNs, in such cases, tend to overfit due to their poor generalizations to large data variances. Data augmentation (DA) is a simple, yet efficient and commonly used strategy to overcome this issue by synthesizing more training examples to artificially increase the dataset~\cite{cubuk2019autoaugment,zhao2019data,zoph2020learning,shen2020anatomical}. It is often employed as a preprocessing step applied to the training data to expose the DNNs to various types of approximated invariance; hence reducing the problem of overfitting. In the image domain, common augmentation strategies include adding variations to image intensities~\cite{chaitanya2019semi,chaitanya2021semi,gao2021enabling,hesse2020intensity,zhao2023augmentation}, or deforming the image by randomly generated geometric transformations (i.e., translation, rotation, or scaling)~\cite{moreno2018forward}. While intensity-based augmentation methods are particularly valuable for tasks that rely heavily on color and texture information, geometric transformations generate diverse, realistic scenarios introducing structural variations in the training data. This diversity in training data leads to improved generalization, as the model learns to recognize key features regardless of their spatial arrangement. In this paper, we focus on DA, particularly in geometric spaces that characterize deformable shape (changes) of objects presented in images~\cite{Joshi2004,beg2005computing,zhang2013bayesian,zhang2015finite,dalca2019learning}. 

In practice, geometric DA is usually a manual process where a human specifies a set of affine or deformable transformations believed to make DL models invariant~\cite{hendrycks2019augmix, yun2019cutmix}. However, such methods can pose a high risk of producing unrealistic or implausible variations of the image. For example, rotating an image 180 degrees makes sense for natural images, but produces ambiguities on digital numbers (i.e., the digits `6’ and `9’). In medical domain, randomly generated deformations may transform images from a healthy to a diseased group. To alleviate this issue, researchers have developed a group of automatic DA methods that learn an optimal augmenting policy from the data itself~\cite{cubuk2019autoaugment,lim2019fast,zoph2020learning,araslanov2021self,muralikrishnan2022glass}.

Despite showing encouraging results on relatively easy datasets (e.g., $28^2$ MNIST), they do not appear to be applicable to more complex datasets with large structural and geometric variances. Later, generative models (e.g., VAE) have been extensively used as a practical and effective solution to capture complex geometric variances for the purpose of DA in low-data regimes~\cite{moreno2020improving,li2019insufficient,pesteie2019adaptive,tang2020onlineaugment}. The rationale for using VAEs in DA lies in their ability to learn and represent latent features of the data. By sampling from the learned latent space, VAEs can generate diverse and realistic synthetic data samples beyond the training dataset. Besides, in the context of DA, VAEs are not typically tasked with learning the entire data distribution but rather with generating synthetic data samples consistent with the original dataset's distribution~\cite{chadebec2021data,chadebec2022data,muralikrishnan2022glass}.

Another common approach is to learn a probabilistic distribution of a set of deformable transformations derived from the training images~\cite{hauberg2016dreaming,krebs2019learning,dalca2019unsupervised,zhao2019data,olut2020adversarial, shen2020anatomical}. However, these methods assume a unimodal distribution of the underlying transformations between images. They do not provide an appropriate way to learn the full space of geometric transformations among the training data. This greatly limits their power when multimodal data that naturally has a distribution of multiple modes, for example, time-series data/longitudinal images~\cite{di2014autism,lamontagne2019oasis}, or images with sub-populations (i.e., population studies that involve healthy vs. disease groups~\cite{jack2008alzheimer,lamontagne2019oasis}).  

In this paper, we present a novel model, MGAug, that for the first time learns augmenting transformations in a {\em multimodal latent space of geometric deformations}. Inspired by works that have utilized mixture models to capture the underlying distribution of deformations from groupwise images, leading to improved analysis for population studies~\cite{zhang2015mixture, zhang2019mixture, luo2020mvmm, hertz2020pointgmm}, our MGAug features the main advantages of
\begin{itemize}
\item Generalizing the automatic DA process into a geometric space that allows hidden transformations with multiple modes.
\item Developing a new generative model that learns the latent distribution of augmented transformations via a mixture of multivariate Gaussians defined as a prior in the tangent space of diffeomorphisms. 
\item Improving the quality of augmented samples from learned knowledge of training data itself with intra-group constraints.
\end{itemize}
To achieve our goal, we first develop a deep learning network that embeds the learning of latent geometric spaces of diffeomorphic transformations via template-based image registration in a VAE. A mixture of multivariate Gaussians will be defined in the tangent space of diffeomorphisms and serve as a prior in the latent space of VAE. We then augment the original training dataset by deforming images using decoded velocity fields that are randomly sampled from the learned multimodal distributions. We demonstrate the effectiveness of our proposed model in two distinct domain-specific tasks: multi-class classification and segmentation on both 2D synthetic data and real 3D brain magnetic resonance images (MRIs), respectively. We also compare MGAug with the state-of-the-art transformation-based augmentation algorithms~\cite{olut2020adversarial,dalca2019unsupervised}. Experimental results show that our method outperforms all baseline algorithms by significantly improved prediction accuracy. 

The remainder of this paper is organized as follows: Sec.~\ref{sec:back} discusses the background of geometric deformations via template-based image registration, Sec.~\ref{subsec:gm} introduces the methodological development of our proposed augmentation algorithm MGAug along with joint optimization scheme, Sec.~\ref{sec:exp} discusses the experimental evaluations that validate the effectiveness of our model on different image analysis tasks, and Sec.~\ref{sec:con} concludes with potential future directions.

\section{Background: Geometric Deformations via Template-based Image Registration}
\label{sec:back}
We first briefly review the concept of template-based deformable image registration~\cite{Joshi2004,zhang2013bayesian}, which is widely used to estimate deformable geometric deformations from groupwise images. In this work, we adopt the algorithm of large deformation diffeomorphic metric mapping (LDDMM), which encourages the smoothness of deformation fields and preserves the topological structures of images~\cite{beg2005computing}.

Given a number of $N$ images $\{I_1,\cdots,I_N\}$, the problem of template-based image registration is to find the shortest paths (a.k.a. geodesics) between a template image $I$ and the rest of image data with derived associate deformation fields $\phi_1,\cdots, \phi_N$ that minimizes the energy function
\begin{equation}
\label{eq:lddmm}
E(I, \phi_n) =\sum_{n=1}^{N} \frac{1}{\sigma^2} \text{Dist} [I \circ \phi_n(v_n(t)), I_n] \, +\, \text{Reg} [\phi_n(v_n(t))],
\end{equation} 
where $\sigma^2$ is a noise variance and $\circ$ denotes an interpolation operator that deforms image $I$ with an estimated transformation $\phi_n$. The $\text{Dist}[\cdot, \cdot]$ is a distance function that measures the dissimilarity between images, i.e., sum-of-squared differences~\cite{beg2005computing}, normalized cross correlation~\cite{avants2008symmetric}, and mutual information~\cite{wells1996multi}. The $\Reg[\cdot]$ is a regularizer that guarantees the smoothness of transformations, i.e.,

\begin{align}
    &\Reg[\phi_n(v_n(t))] = \int_0^1 (L v_n(t), v_n(t)) \, dt, \nonumber \\ 
    &\quad \text{with} \quad \frac{d\phi_n(t)}{dt} =  v_n(t) \circ \phi_n(t),   
    \label{eq:distance}
\end{align}
where $L: V\rightarrow V^{*}$ is a symmetric, positive-definite differential operator that maps a tangent vector $ v_n(t)\in V$ into its dual space as 
a momentum vector $m_n(t) \in V^*$. We write $m_n(t) = L v_n(t)$, or $v_n(t) = K m_n(t)$, with $K$ being an inverse operator of $L$. The operator $D$ denotes a Jacobian matrix and $\cdot$ represents element-wise matrix multiplication. In this paper, we use a metric of the form $L = -\alpha \Delta + \mathbb{I}$, in which $\Delta$ is the discrete Laplacian operator, $\alpha$ is a positive regularity parameter, and $\mathbb{I}$ denotes an identity matrix.

The minimum of Eq.~\eqref{eq:distance} is uniquely determined by solving a Euler-Poincar\'{e} differential equation (EPDiff)~\cite{arnold1966geometrie,miller2006geodesic} with a given initial condition of velocity fields, noted as $v_n(0)$,
\begin{align}
    \frac{\partial v_n(t)}{\partial t} &= - K [(D v_n(t))^T \cdot m_n(t) + D m_n(t) \cdot v_n(t) \nonumber \\ &+ m_n(t) \cdot \operatorname{div} v_n(t) ],
    \label{eq:epdiff}
\end{align}
where the operator $D$ denotes a Jacobian matrix, $\operatorname{div}$ is the divergence, and $\cdot$ represents an element-wise matrix multiplication. This is known as the geodesic shooting algorithm~\cite{vialard2012diffeomorphic}, which nicely proves that the deformation-based shape descriptor, $\phi_n$, can be fully characterized by an initial velocity field, $v_n(0)$.

We are now able to equivalently minimize the energy function in Eq.~\eqref{eq:lddmm} as
\begin{align}
   E(I, v_n(0)) &=  \sum_{n=1}^{N} \frac{1}{\sigma^2} \text{Dist} [I \circ \phi_n(v_n(0)), I_n] \nonumber\\ &+ (L v_n(0), v_n(0)), \, \,  \text{s.t. Eq.~\eqref{eq:epdiff}.}   
\label{eq:flddmm}
\end{align}
For notation simplicity, we will drop the time index in the following sections.

Numerous research works have defined a prior on the initial velocity field, $\tilde{v}_n$, as a multivariate Gaussian distribution to ensure the smoothness of the geodesic path~\cite{wang2021bayesian,zhang2013bayesian,zhang2016low}. This prior is typically formulated as 
\begin{align}
\label{eq:prior}
p({v}_n) = \frac{1}{(2 \pi)^{\frac{M}{2}} | {L}^{-1} |^{\frac{1}{2}}} \exp{ \left(-\frac{1}{2}({L} {v}_n, {v}_n) \right)},
\end{align}
where $M$ denotes the total number of voxels of initial velocities, and $|\cdot|$ is the matrix determinant. Note that here we consider the input images to be measured on a discretized grid. That is, images are elements of the finite-dimensional Euclidean space. We will also consider velocity fields and the resulting diffeomorphisms to be defined on the discrete grid. This allows us to work with a well-defined probability density function while maintaining the smoothness properties of the velocity fields through the use of the differential operator $L$. In practice, we employ a discretized operator $L$ (as previously defined in Eq.~\eqref{eq:distance})~\cite{beg2005computing,vialard2012diffeomorphic,zhang2013bayesian}. The $|L^{-1}|$ refers to the determinant of the inverse of the discretized differential operator.

\section{Our Method: MGAug}
\label{sec:method}
This section presents a novel algorithm, MGAug, that learns a multimodal latent distribution of geometric transformations represented by the initial velocity fields $v_0$. We first introduce a generative model in latent geometric spaces for DA based on VAE, followed by an approximated variational inference. The augmented data is then used to train various image analysis tasks, i.e., image classification or segmentation. Similar to~\cite{zhao2019data, chen2022enhancing}, we employ a joint learning scheme to optimize the MGAug and image analysis tasks (IAT) network parameters. To simplify the notation, we omit the time index of  $v_0$ in the following sections. 

\subsection{Generative Model in Latent Multimodal Geometric Spaces}
\label{subsec:gm}
\begin{figure}[htb]
   \centering
   \includegraphics[width=0.7\linewidth]{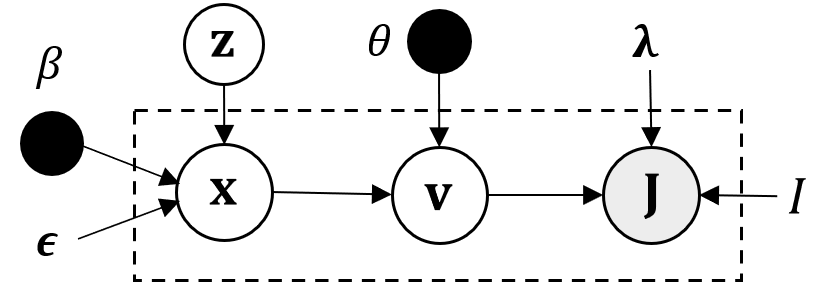}
   \caption{Graphical representation of our proposed generative model in multimodal latent space. }
    \label{fig:graphicalmodel}
\end{figure}

Given a group of $N$ input images $\mathbf{J}$ with a template $I$, we assume the latent geometric transformations $\mathbf{x}$ are generated from a number of $C$ clusters. The distribution of $\mathbf{x}$ can be formulated as a Gaussian mixture with means $\boldsymbol \mu$ and variances $\mathbf{\Sigma}$, specified by a VAE encoder parameterized by $\beta$, i.e., $p_{\theta}(\mathbf{x}|{\boldsymbol\epsilon})$, where $\boldsymbol\epsilon$ is a random number generated from $\mathcal{N}(0, \mathbb{I})$. That is to say, the VAE encoder outputs a set of $C$ means $\boldsymbol\mu$ and $C$ variances $\mathbf{\Sigma}$. A one-hot vector $\mathbf{z}$ is sampled from the mixing probability $\mathbf{\boldsymbol\pi}\in [0, 1]$ with $\sum_{c=1}^C \mathbf{\boldsymbol\pi}^c = 1$, which chooses one component from the Gaussian mixture. In this paper, we set the parameter $\mathbf{\boldsymbol\pi} = C^{-1}$ to make the variable $\mathbf{z}$ uniformly distributed, which is similar to~\cite{dilokthanakul2016deep}. The initial velocity field $\mathbf{v}$ is generated from a VAE decoder parameterized by $\theta$ and the continuous latent variable $\mathbf{x}$. The entire process to generate the observed image data $\mathbf{J}$ from a set of latent variables $\{\mathbf{v},\mathbf{x}, \mathbf{\boldsymbol\epsilon}, \mathbf{z}\}$ is:
\begin{align}
\label{eq:epsilon} 
\mathbf{\boldsymbol\epsilon} &\sim \mathcal{N}(0, \mathbb{I}),\\
\mathbf {z} &\sim \text{Mult}(\mathbf{\boldsymbol\pi}),  \\
\mathbf{x} | \mathbf{z}, \mathbf {\boldsymbol\epsilon} &\sim \prod_{c=1}^C \mathcal{N}(\mathbf{\boldsymbol\mu}_{\mathbf{z}_c}(\mathbf{\boldsymbol\epsilon}, \beta), \mathbf{\Sigma}_{\mathbf{z}_c}(\mathbf{\boldsymbol\epsilon}, \beta))^{\mathbf{z}_c},  \\
\mathbf{v} | \mathbf{x} &\sim \mathcal{N}(\mathbf{v}_\theta(\mathbf{x}) \, | \, 0, \mathbf{\Sigma}) \label{eq:mulGD}, \\
\mathbf{J} | \mathbf{v}, I, \mathbf{\lambda} &\sim \mathcal{N}(\mathbf{J} \, | \, I \circ \mathbf{\phi}(\mathbf{v}), \boldsymbol\lambda^2 \mathbb{I}), \label{eq:ID_like}
\end{align}
where $\text{Mult}(\cdot)$ denotes a multinomial distribution, $\mathbf{\boldsymbol\lambda}^2$ is a noise variance in the observed image space, and an identity matrix $\mathbb{I}$ maps a matrix to be diagonal. 
A graphical representation of our proposed generative model is shown in Fig.~\ref{fig:graphicalmodel}. 

To guarantee the smoothness of the geometric transformations, we define a multivariate Gaussian distribution on the generated initial velocity fields $\mathbf{v}$ in Eq. (\ref{eq:mulGD}) as
\begin{equation}
\label{eq:priorv}
p(\mathbf{v}|\mathbf{x}) \propto \exp \left(-\frac{1}{2}(\mathbf{\Sigma}^{-1} \mathbf{v}_\theta(\mathbf{x}), \mathbf{v}_\theta(\mathbf{x})) \right).
\end{equation}
Similar to Eq.~\eqref{eq:prior}, we define the covariance matrix $\boldsymbol\Sigma$ as an inverse operator of the discretized differential operator, $L$, which is typically used in the registration regularity in Eq.~\eqref{eq:distance} to guarantee the smoothness of transformations and preserve the topological structures of images~\cite{beg2005computing}.

Considering that our input images are measured on a discrete grid, we formulate our image noise model as an i.i.d. Gaussian at each image voxel, i.e.,
\begin{align}
\label{eq:likelihoodv}
p(\mathbf{J} | \mathbf{v}, I, \mathbf{\boldsymbol\lambda}) = \frac{1}{(2 \pi)^{M/2} (\boldsymbol\lambda)^M} \exp\left(-\frac{\|I \circ \mathbf{\phi}(\mathbf{v})-\mathbf{J}\|_{2}^2}{2 \boldsymbol\lambda^2}\right),
\end{align}
where $M$ is the total number of image voxels. Putting together Eq. (\ref{eq:epsilon}) to Eq.~\eqref{eq:likelihoodv}, we arrive at a generative model 
\begin{equation}
\label{eq:gm}
p_{\beta, \theta}(\mathbf{J}, \mathbf{v}, \mathbf{x}, \mathbf{\boldsymbol\epsilon}, \mathbf{z})=p(\mathbf{J} | \mathbf{v}, I, \mathbf{\boldsymbol\lambda})\,p(\mathbf{v} | \mathbf{x})\,p(\mathbf {x} | \mathbf{z}, \mathbf {\boldsymbol\epsilon})\,p(\mathbf {z})\,p(\mathbf{\boldsymbol\epsilon})
\end{equation}

\subsection{Variational Lower Bound}
\label{subsec:vlb}
We will train the generative model in Eq.~\eqref{eq:gm} with the variational inference objective, i.e. the log-evidence lower
bound (ELBO) \cite{kingma2013auto}. With an assumption that the mean-field variational family $q(\mathbf{v}, \mathbf{x}, \mathbf{\boldsymbol\epsilon}, \mathbf{z} \, | \, \mathbf{J})$ is an approximate to the posterior, we are able to formulate the ELBO as
\begin{equation}
\mathcal{L}_{\text{ELBO}}=\mathbb{E}\left[\log\frac{p_{\beta, \theta}(\mathbf{J}, \mathbf{v}, \mathbf{x}, \mathbf{\boldsymbol\epsilon}, \mathbf{z})}{q(\mathbf{v}, \mathbf{x}, \mathbf{\boldsymbol\epsilon}, \mathbf{z} \, | \, \mathbf{J})}\right]\\
\end{equation}

We factorize the mean-field variational family as 

\begin{equation}
\mathbf{q} = \prod_{i=1}^n q(\mathbf{v}_i|\mathbf{J}_i, I, \boldsymbol\lambda)\, q_{\psi_\mathbf{x}}(\mathbf{x}_i|\mathbf{v}_i, \mathbf{J}_i)\,q_{\psi_\mathbf{\boldsymbol\epsilon}}(\mathbf{\boldsymbol\epsilon}_i|\mathbf{v}_i)\,p_{\beta}(\mathbf{z}_i|\mathbf{x}_i, \mathbf{\boldsymbol\epsilon}_i),
\end{equation}

where $i$ denotes the $i$-th image. Each variational factor is parameterized with the recognition networks $\psi_\mathbf{x}$ and $\psi_\mathbf{\boldsymbol\epsilon}$ that output the parameters of the variational Gaussian distributions. Note that we will drop the index $i$ in the following sections for the purpose of notation simplicity.\\

\begin{figure*}[!t]
    \centering
    \includegraphics[width=\textwidth] {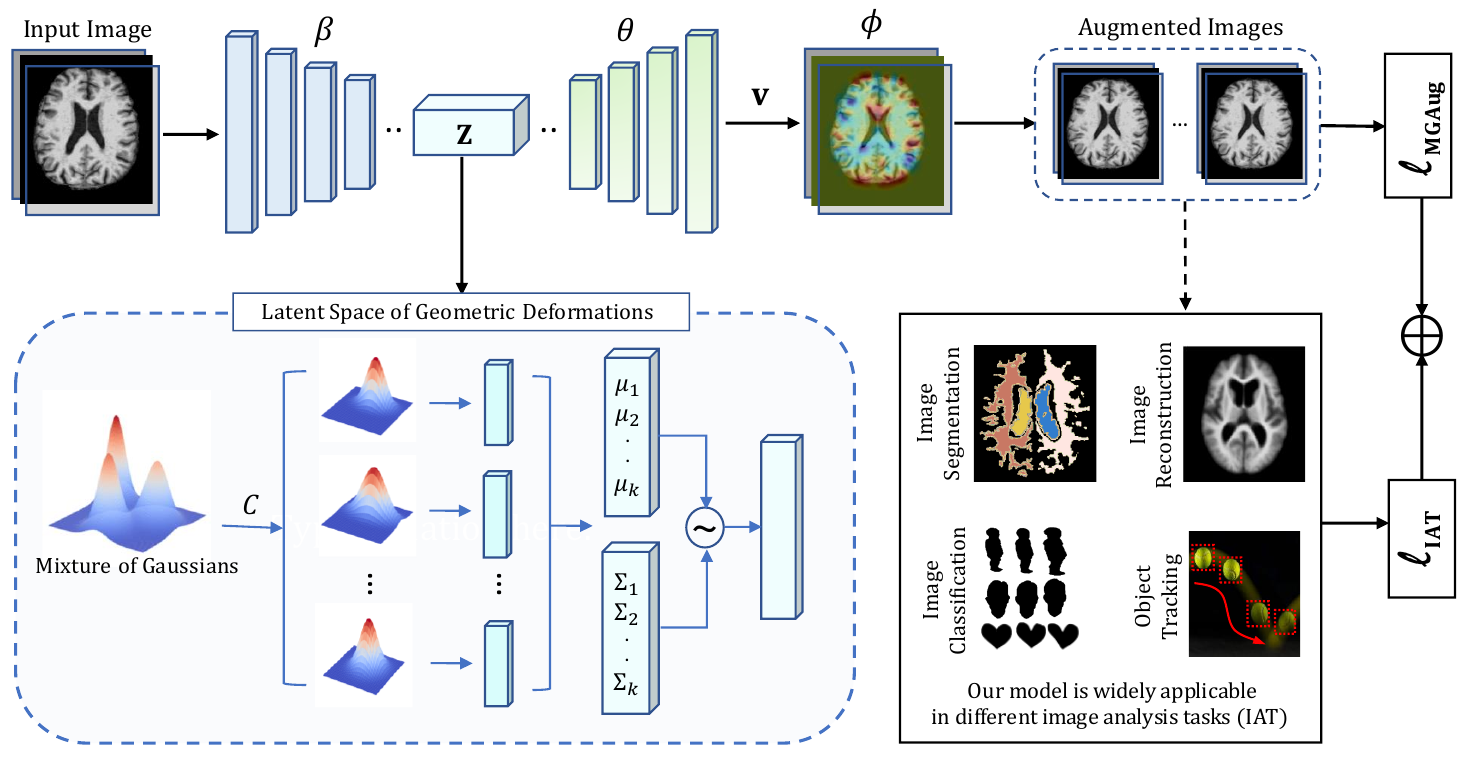}
    \caption{An overview of our model MGAug for image analysis tasks.}
    \label{fig:MGAug}
\end{figure*}

Using previously defined Eq.~\eqref{eq:priorv} and Eq.~\eqref{eq:likelihoodv}, the posterior $q(\mathbf{v}_i|\mathbf{J}_i, I, \boldsymbol\lambda) \propto p(\mathbf{J} | \mathbf{v}, I, \mathbf{\boldsymbol\lambda}) \cdot p(\mathbf{v}|\mathbf{x})$ is formulated by Baye's rule. Following~\cite{jiang2016variational}, we can derive the posterior $p_{\beta}(\mathbf{z}|\mathbf{x}, \mathbf{\boldsymbol\epsilon})$ as
\begin{align*}
\centering
    p_{\beta}(\mathbf{z}_{j}=1|\mathbf{x}, \mathbf{\boldsymbol\epsilon}) &= \frac{p(\mathbf{z}_{j}=1)\,p(\mathbf{x}|\mathbf{z}_{j}=1,\boldsymbol\epsilon)}{\sum_{c=1}^{C}p(\mathbf{z}_{c}=1)\,p(\mathbf{x}|\mathbf{z}_{j}=1, \boldsymbol\epsilon)}\nonumber\\
    &= \frac{\boldsymbol\pi_{j}\, \mathcal{N}(\mathbf{x}|\boldsymbol\mu_{j}(\boldsymbol\epsilon;\beta), \boldsymbol\Sigma_{j}(\boldsymbol\epsilon; \beta))}{\sum_{c=1}^{C}\boldsymbol\pi_{\mathbf{z}_c}\, \mathcal{N}(\mathbf{x}|\mu_{\mathbf{z}_c}(\boldsymbol\epsilon; \beta), \Sigma_{\mathbf{z}_c}(\boldsymbol\epsilon; \beta))},
\end{align*}
where $\mathbf{z}_j, j \in \{1, \cdots, C\}$ denotes the $j$-th element of $\mathbf{z}$. The variational lower bound can be now formulated as
\begin{align}
    \mathcal{L}_{ELBO} &= \mathbb{E}_{q(\mathbf{x}|\mathbf{J})}[\log P_{\theta}(\mathbf{J}|\mathbf{x})] \nonumber \\ &- \mathbb{E}_{q(\boldsymbol\epsilon|\mathbf{J})\,p(\mathbf{z}|\mathbf{x},\boldsymbol\epsilon)}[KL(q_{\psi_{\mathbf{x}}}(\mathbf{x}|\mathbf{J})\,||\,p_{\beta}(\mathbf{x}|\boldsymbol\epsilon, \mathbf{z}))] \nonumber\\
    &- KL(q_{\psi_{\mathbf{\boldsymbol\epsilon}}}(\boldsymbol\epsilon|\mathbf{J})\,||\,p(\boldsymbol\epsilon)) \nonumber \nonumber \\ &- \mathbb{E}_{q(\mathbf{x}|\mathbf{J})\,q(\boldsymbol\epsilon|\mathbf{J})}[KL(p_{\beta}(\mathbf{z}|\mathbf{x}, \boldsymbol\epsilon)\,||\,p(\mathbf{z}))] \label{eq:elbo_rec}
\end{align}
\label{subsec:nl}

\subsection{Joint Learning of MGAug with Image Analysis}
We demonstrate the effectiveness of MGAug in image analysis tasks (IAT) by developing a joint learning framework. It has been shown in~\cite{zhao2019data,chen2022enhancing} that these two tasks can be mutually beneficial. The MGAug effectively synthesizes image samples to augment the training dataset, while the IAT can provide extra information on rectifying the learned distributions of latent geometric spaces. An overview of our proposed network architecture is shown in Fig.~\ref{fig:MGAug}.

\subsubsection{Network design of MGAug} 
Our MGAug is mainly designed based on the previously introduced inference model, where the approximate posteriors are simulated by neural networks. Analogous to~\cite{kingma2013auto}, we let the encoder of MGAug, parameterized by $\beta$, approximate the posterior as a mixture of multivariate Gaussians. In the latent space $\mathbf{x}$, we use non-linear hidden layers to learn the parameters (i.e., means and covariances) of the mixture Gaussians. The decoder, parameterized by $\theta$, outputs augmented velocity fields $\mathbf{v}_{n}$, followed by generating their associated deformations $\phi_n$ (via Eq.~\eqref{eq:epdiff}) and augmented deformed images. In this paper, we adopt UNet architecture~\cite{ronneberger2015u} as the network backbone for MGAug. However, other networks such as UNet++~\cite{zhou2018unet++} and TransUNet~\cite{chen2021transunet} can be easily applied.\\ 

Defining $\Theta = (\mathbf{J}, \beta, \theta,  \mathbf{x}, \mathbf{z}, \mathbf{v}, \boldsymbol\epsilon, I)$ as posterior parameters, we formulate the network loss of MGAug as a combination of dissimilarity between deformed templates and input images, a regularity to ensure the smoothness of augmenting transformations, and the variational inference objective as
\begin{equation}
\begin{aligned}
    \ell_{\text{MGAug}}(\Theta) &= -\log p_{\beta, \theta}(\mathbf{J}, \mathbf{v}, \mathbf{x}, \mathbf{\boldsymbol\epsilon}, \mathbf{z})\\&  \approx  \frac{\|I \circ \mathbf{\phi}(\mathbf{v}(\beta, \theta))-\mathbf{J}\|_{2}^2}{2 \boldsymbol\lambda^2} + \mathcal{L}_{ELBO} \nonumber \\ &+ \frac{1}{2}(L \mathbf{v} (\beta, \theta), \mathbf{v} (\beta, \theta)) + \text{reg}(\beta, \theta),
    \label{loss:maguag}
\end{aligned}
\end{equation}
where $\text{reg}(\cdot)$ denotes network parameter regularization.

\subsubsection{Network design of IAT} 
We demonstrate MGAug on the task of image classification and segmentation separately. Other domain-specific tasks, such as image reconstruction, object recognition, and tracking can be integrated similarly.  

For classification, assuming we have augmented images (consist of $C$ classes with $N_{c}$ images per class) and their labels denoted as $\hat{y}$. The classifier is trained by minimizing the loss
\begin{equation}
    \ell_{\text{IAT}}^{clf} = -\gamma \sum_{n=1}^{N_{c}}\sum_{c=1}^{C} \hat{y}_{nc} \cdot \log [ y_{nc} (\beta, \theta, w_1)] + \text{reg}(\beta, \theta, w_1),
    \label{loss:clf}
\end{equation}
where  $y_{nc} $ denotes the predicted label,  $\gamma$ is the weighting parameter, and $w_1$ is a classification network parameter.

For segmentation, we employ a loss function as a combination of S\o rensen$-$Dice coefficient~\cite{dice1945measures} and cross-entropy (CE). Let $Q$ denote the number of segmentation labels, $S_{q}$ denotes the augmented ground truth, and $R_{q}$ denotes the prediction with the associated probabilistic map $R^{p}_{q}$ for $q$-th label.  We formulate the segmentation loss as
\begin{align}
    \ell_{\text{IAT}}^{seg} &= \tau \sum_{q=1}^{Q} [1-\text{Dice} (S_q, R_q (\beta, \theta, w_2)) \nonumber \\
    &+ \text{CE} (S_q, R^p_q (\beta, \theta, w_2))] + \text {reg} (\beta, \theta, w_2), 
    \label{loss:seg}
\end{align}
where $\text{Dice}(S_q, R_q) = 2(|S_q| \cap |R_q|)/(|S_q| + |R_q|) $ and $\text{CE} (S_q, R^p_q )= -S_{q}\cdot\log R_{q}$. The $\tau$ is the weighting parameter and $w_2$ denotes the segmentation network parameters.

The total loss of our joint learning framework of MGAug and IAT is $\ell =\ell_{\text{MGAug}} + \ell_{\text{IAT}}$. We employ an alternative optimization scheme~\cite{nocedal1999numerical} to minimize the loss by alternating between the training of the MGAug network and the IAT network without sharing weights, making it an end-to-end network architecture~\cite{ zhao2019data, chen2022enhancing}.

\begin{algorithm}[!ht]
\SetAlgoLined
\SetArgSty{textnormal}
\SetKwInOut{Input}{Input}
\SetKwInOut{Output}{Output}
  \Input{A group of $N$ input images with an image template $I$, class labels $\hat{y}_{\mathbf{J}}$ or ground truth segmentation labels $\hat{Y}_{\mathbf{J}}$ and a number of iteration $r$.}
  \Output{Predicted classification labels $y_{\mathbf{J}}$ or segmentations $Y_{\mathbf{J}}$.}
  \Repeat{\text{convergence,} $|\Delta \ell| < \epsilon$}{
   \For{i = 1 to $r$}
   {
   \tcc{Train MGAug}
   Minimize the augmentation loss in Eq.~\eqref{loss:maguag}
   
   Generate augmented images (and propagate associated deformations to segmentation labels, for segmentation);
   
   \tcc{Train classifier or segmentation network}
   
   Mix augmented images and labels with ground truth labels, $\hat{y}_{\mathbf{J}}$ or $\hat{Y}_{\mathbf{J}}$, in training;
   
   Minimize the classification loss in Eq.~\eqref{loss:clf} or segmentation loss in Eq.~\eqref{loss:seg};
   
   Output the predicted class labels $y_{\mathbf{J}}$ or segmentations $Y_{\mathbf{J}}$;
   }
}
\caption{Joint Training of MGAug with IAT.}
\label{alg:altopt}
\end{algorithm}

A summary of our joint training of MGAug with IAT is presented in Alg.~\ref{alg:altopt}.

\section{Experimental Evaluation}
\label{sec:exp}

\subsection{Datasets}
We validate our proposed model, MGAug, on both 2D synthetic datasets and real 3D brain MRI scans, highlighting its effectiveness on training data with limited availability.

\paragraph*{\bf 2D silhouette shapes} We choose Kimia216~\cite{sebastian2004recognition}, which is a relatively smaller dataset with seven classes of 2D silhouette shape data (\textit{bone, bottle, children, face, fish, heart}, and \textit{key}). Each shape class includes approximately $20$ binary images, with the size of $28^2$. The total number of our training images for multi-classification is about $140$. 

\paragraph*{\bf 2D handwritten digits} We use $70000$ handwritten digits of $10$ classes ($0$-$9$) from MNIST repository~\cite{lecun1998mnist}. Each class includes approximately $7000$ images with the size of $28^2$. 

\paragraph*{\bf 3D brain MRIs} We include $3338$ publicly available T1-weighted 3D brain MRIs from the Alzheimer's Disease Neuroimaging Initiative (ADNI)~\cite{jack2008alzheimer}.All subjects ranged in age from $50$ to $100$, covering cognitively normals (CN), patients affected by Alzheimer's disease (AD),  individuals with mild cognitive impairment MCI, respectively. All MRIs were preprocessed to be the size of $128\times128\times128$, and underwent skull-stripping, intensity min-max normalization, bias-field correction, and affine registration~\cite{reuter2012within}.

\paragraph*{\bf 3D brain MRIs w/ segmentations} We include $803$ subjects of public T1-weighted longitudinal brain MRIs with manually delineated anatomical structures from Open Access Series of Imaging Studies (OASIS)~\cite{lamontagne2019oasis}, Alzheimer's Disease Neuroimaging Initiative (ADNI)~\cite{jack2008alzheimer}, Autism Brain Imaging Data Exchange (ABIDE)~\cite{di2014autism}, and LONI Probabilistic Brain Atlas Individual Subject Data (LPBA40)~\cite{shattuck2008construction}. Due to the difficulty of preserving the diffeomorphic property across individual subjects with large age variations, we carefully evaluate images from subjects aged from $60$ to $90$. All MRIs were pre-processed as $256 \times 256 \times 256$, $1.25mm^{3}$ isotropic voxels, and underwent skull-stripping, intensity normalization, bias field correction and pre-alignment with affine transformations.

\subsection{Experiments and Implementation Details}
We validate the proposed model, MGAug, on both image classification and segmentation tasks. As MGAug falls into the category of transformation-based DA schemes, we mainly focus on comparing its performance with similar baselines: randomly generated affine transformations (\textit{ATA}, a commonly used augmentation strategy), and two state-of-the-art unimodal generative models based on diffeomorphic registrations (\textit{AdvEigAug or AEA}, an augmentation model based on the statistics of a learned latent space of velocity fields~\cite{olut2020adversarial}; and \textit{UDA}, a unimodal deformation-based augmentation using random stationary velocity fields~\cite{dalca2019unsupervised}). 

\paragraph*{\bf Evaluation of classification on 2D datasets} We first validate our model and other baselines on a relatively small dataset of 7-class 2D silhouette shapes. We adopt six classification network backbones, including a three-layer CNN, ResNet18~\cite{he2016deep}, VGGNet16~\cite{simonyan2014very}, SqueezeNet~\cite{iandola2016SqueezeNet}, DenseNet121~\cite{huang2017densely}, and Vision Transformer (ViT)~\cite{dosovitskiy2020image}. 

We carefully determine the optimal number of modes, $C$, for our multimodal generative model by performing cross-validation with various values of $C$. The value of \textit{C} will vary for different image analysis tasks depending on the dataset distributions. For classification, we assume that the value of \textit{C} is close to the number of classes. Furthermore, we conduct a comprehensive assessment of the augmentation efficiency offered by our proposed approach. This involves augmenting the training dataset to various magnitudes, ranging from $2$ to $10$ times the size of the original dataset. We further reduce the number of ground truth images to $75\%$ and $50\%$ to experiment with the effectiveness of the augmented data on this multi-class classification task.

To examine the quality of the generated augmenting transformations, we compute the determinant of Jacobian of the deformation fields. Throughout the evaluation process, we employ multiple metrics such as average accuracy (Acc), precision (Prec), and F1-score (F1) to validate the classification performance. 

We further validate the effectiveness of our augmentation model by experimenting on a relatively larger handwritten mnist dataset. As classification on mnist using all $75000$ ground truths almost achieves perfect classification accuracy, we experiment with $10\%$ ground truths. First, we perform classification under MLP ($4$ linear layers) and $2$-block CNN over an increased number of augmented data from $1\times$ to $5\times$. Each CNN block is built with one $3\times3$ convolutional layer, one $2\times2$ pooling layer, and two linear layers with ReLU activations. Second, we further validate our augmentation model by comparing its performance across a wider range of network backbones, including ResNet18/50~\cite{he2016deep}, VGGNet16/19~\cite{simonyan2014very}, DenseNet121/169~\cite{huang2017densely}.

\paragraph*{\bf Evaluation of classification on real 3D brain MRIs} We evaluate the efficiency of MGAug on real-world 3D brain images, where we identify whether the subject is healthy, diagnosed with Alzheimer's, or early/late mild cognitive impairment. Similar to the experiential design of 2D data, we conduct comprehensive experiments of the augmentation efficiency of MGAug, augmenting the training dataset to various magnitudes ($1\times$ to $5\times$) over different ground-truth settings ($25\%$, $50\%$, and $75\%$). Additionally, we examine the network performance on various numbers of modes $C$.

\paragraph*{\bf Evaluation of segmentation on real 3D brain MRIs} We validate the performance of all methods on real 3D brain images with a decent size of training data. In this experiment, we employ UNet~\cite{ronneberger2015u}, UNetR~\cite{hatamizadeh2022unetr}, TransUNet~\cite{chen2021transunet} as our network backbone, given its proven superiority in brain segmentation performance. Our goal is to demonstrate that our proposed multimodal distribution learning of the deformable geometric latent space will improve the segmentation performance, regardless of what particular network backbone is. To ensure a fair comparison of the final segmentation accuracy, we augment the training data three times for all methods, resulting in a total number of approximately $2000$ augmented images. We then compare the segmentation dice scores of all augmentation methods over different brain structures.

\paragraph*{\bf Evaluation of the quality of augmented images} To further evaluate the quality of the augmented images, we validate MGAug on two relatively large datasets (2D MNIST and 3D brain MRIs) with various levels of reduced training sample sizes (i.e., $10\%$, $20\%$, and $50\%$ of the original size). We augment the training images three times of the original size for all methods. 

\paragraph*{\bf Comparison with random/adversarial augmentation} We further compare our augmentation model with two additional deformable augmentations: random deformable augmentations and adversarial data augmentation \cite{chen2022enhancing}. The purpose of this comparison is to highlight the effectiveness of MGAug in generating plausible and structurally consistent deformations while maintaining superior performance in preserving the integrity of medical images and improving downstream tasks, including classification and segmentations.

\paragraph*{\bf Statistical evaluation} We perform paired t-tests~\cite{student1908probable} and report the corresponding P-values for all models across both classification and segmentation tasks on real 3D brain MRI datasets. This analysis provides insights on whether MGAug performs significantly better or differently compared to baselines.

\paragraph*{\bf Parameter setting} We set parameter $\alpha=3.0$ for the operator $\mathcal{L}$ and the number of time steps for Euler integration in EPDiff (Eq.~\eqref{eq:epdiff}) as $10$. We set the noise variance $\sigma$ as $0.02$, the batch size as $16$, and use the cosine annealing learning rate scheduler that starts from a learning rate $\eta = 1e^{-3}$ for network training.  We run all the models on training and validation images with Adam optimizer until obtaining the best validation performance. The training and prediction procedure of all learning-based methods are performed on a 40GB NVIDIA A100 Tensor Core GPU. For both 2D and 3D datasets, we split the images by using $70\%$ as training, $15\%$ as validation, and $15\%$ as testing images. 

\paragraph*{\bf Template selection} Our method is flexible to either automatically estimated or pre-selected template(s) for group-wise data. For the classification tasks involving 2D shapes/images and 3D brain MRI datasets, we pre-selected class-specific templates that are most representative, aligning the number of templates with the number of classes. This choice is motivated by the principle that augmented deformations should preserve group-wise structural properties, ensuring that class-specific templates maintain the topological integrity necessary for accurate classification. For segmentation tasks, we pre-selected three templates representing the healthy, diseased, and mild cognitive impairment classes. Please note that automatically estimating template images could be an alternative in our model.

In all experiments presented in this paper, we run cross-validation to determine the optimal number of modes $C$.

\section{Results}

\begin{table*}[!htb]
  \caption{Accuracy comparison on 2D shapes over all models with various network backbones.}
  \centering
  \resizebox{\textwidth}{!}{\begin{tabular}{lcccccccccccc}
  \toprule
    & \multicolumn{3}{c}{\textbf{ATA}}                         & \multicolumn{3}{c}{\textbf{AEA}}                        & \multicolumn{3}{c}{\textbf{UDA}}                         & \multicolumn{3}{c}{\textbf{MGAug}}       \\
    & \textit{Acc} & \textit{Prec} & \textit{F1}                & \textit{Acc} & \textit{Prec} & \textit{F1}                & \textit{Acc} & \textit{Prec} & \textit{F1}                & \textit{Acc} & \textit{Prec} & \textit{F1} \\
                                  \toprule
\multicolumn{1}{l|}{CNN}    & 71.73        & 73.25        & \multicolumn{1}{c|}{66.89} & 83.93        & 88.82        & \multicolumn{1}{c|}{83.07} & 85.71        & 87.14        & \multicolumn{1}{c|}{83.73} & \textbf{91.07}        & \textbf{92.14}        & \textbf{90.78}      \\
\multicolumn{1}{l|}{ResNet18}    & 76.78        & 79.22        & \multicolumn{1}{c|}{73.44} & 80.36        & 82.17        & \multicolumn{1}{c|}{78.76} & 87.50        & 90.07        & \multicolumn{1}{c|}{86.83} & \textbf{89.28}        & \textbf{90.43}        & \textbf{89.15}       \\

\multicolumn{1}{l|}{VGGNet16}    & 73.21        & 72.46        & \multicolumn{1}{c|}{71.23} & 85.71        & 88.04        & \multicolumn{1}{c|}{85.85} & 85.71        & 88.57        & \multicolumn{1}{c|}{81.58} & \textbf{89.28}        & \textbf{91.21}        & \textbf{89.22}       \\
\multicolumn{1}{l|}{SqueezeNet}  & 75.00        & 75.85        & \multicolumn{1}{c|}{72.99} & 76.78        & 79.90        & \multicolumn{1}{c|}{74.53} & 83.93        & 87.30        & \multicolumn{1}{c|}{82.73} & \textbf{92.86}        & \textbf{93.97}        & \textbf{92.59}       \\
\multicolumn{1}{l|}{DenseNet121} & 75.89        & 77.54        & \multicolumn{1}{c|}{73.88} & 87.50         & 90.79        & \multicolumn{1}{c|}{86.18} & 89.28 & 90.86        & \multicolumn{1}{c|}{89.08} & \textbf{94.64}        & \textbf{95.55}        & \textbf{94.28}      \\

\multicolumn{1}{l|}{ViT} & 78.57        &  82.62        & \multicolumn{1}{c|}{76.53} & 89.28         &    92.20     & \multicolumn{1}{c|}{89.33} & 91.07        & 91.93        & \multicolumn{1}{c|}{96.84} & \textbf{98.21}        & \textbf{98.41}        & \textbf{98.20}      \\
\bottomrule
\end{tabular}}
\label{table:clf_SOTA_comp}
\end{table*}

Tab.~\ref{table:clf_SOTA_comp} reports the multi-class classification accuracy on a relatively smaller 2D silhouette shape data across all augmentation methods. For a fair comparison, we augment the data $10\times$ of the original training images. MGAug achieves the highest performance, outperforming all baselines by $2$-$9\%$ of accuracy on six different network backbones. This further proves that our augmented images help classifiers of any size (small: CNN to large: ViT) to achieve optimal performance.

\begin{figure}[!htb]
\centering
\begin{subfigure}{0.45\linewidth}  
  \centering
  \includegraphics[width=\linewidth]{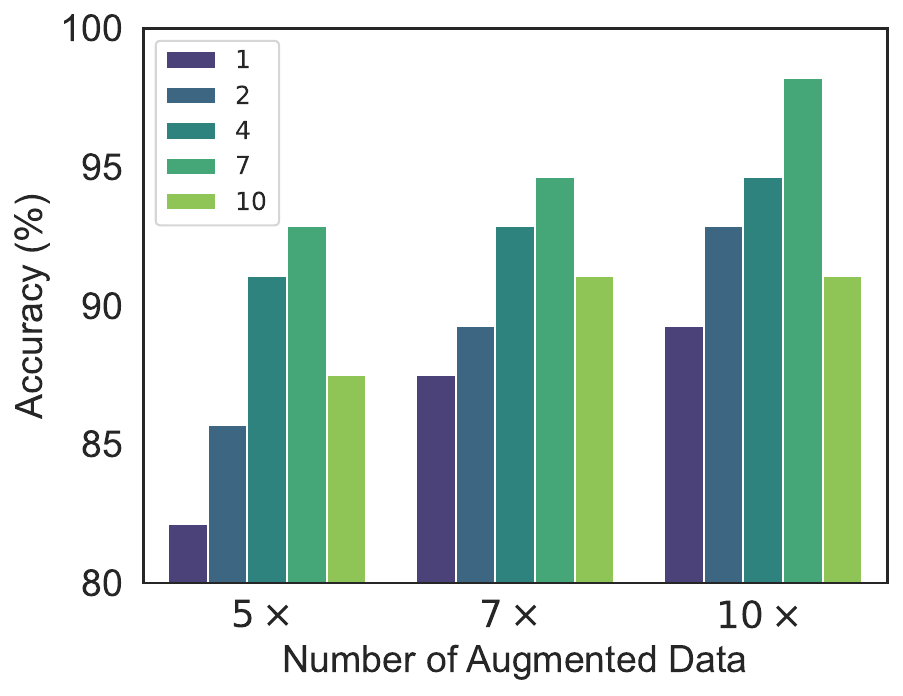} 
\end{subfigure}
\begin{subfigure}{0.45\linewidth}  
  \centering
  \includegraphics[width=\linewidth]{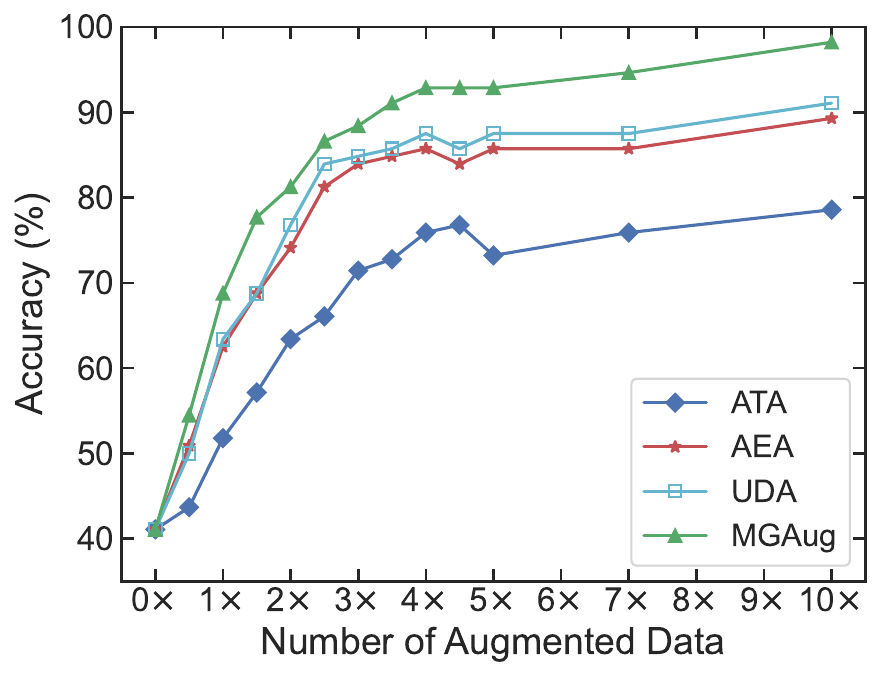} 
\end{subfigure}
\par\vspace{0.5cm}  
\begin{subfigure}{0.45\linewidth}  
  \centering
  \includegraphics[width=\linewidth]{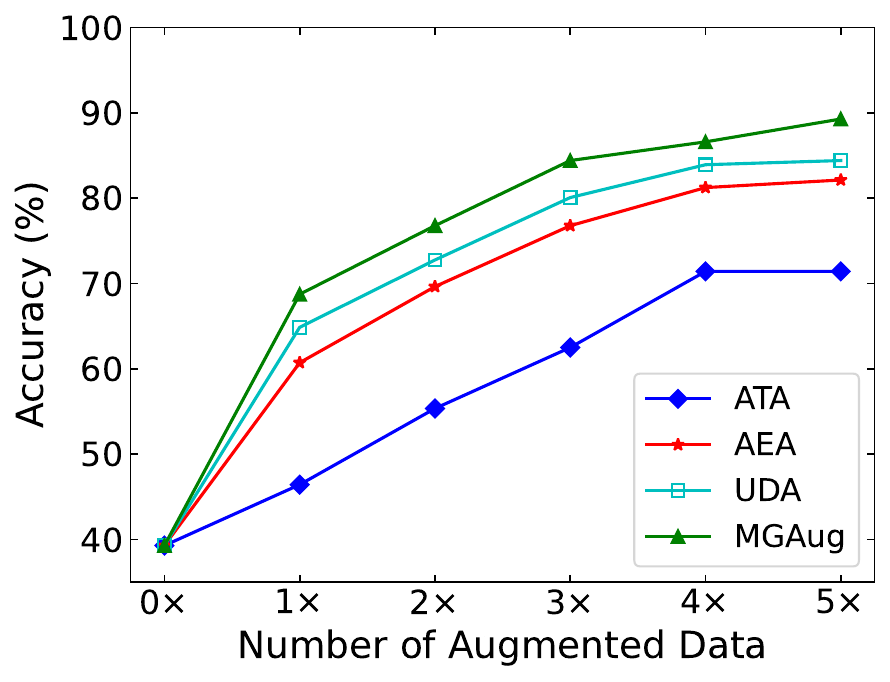} 
\end{subfigure}
\begin{subfigure}{0.45\linewidth}  
  \centering
  \includegraphics[width=\linewidth]{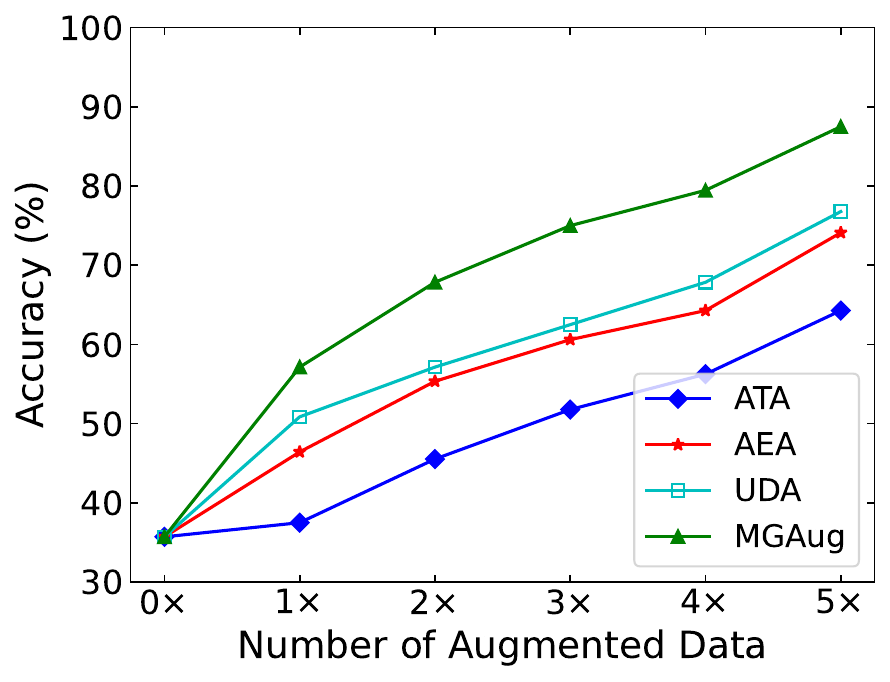} 
\end{subfigure}
\caption{Left: Accuracy comparison on various modes $C$ on 2D shape dataset. Right: A comparison of classification performance for all models over increasing number of augmented data taking all (top-right), $75\%$ (bottom-left), and $50\%$ (bottom-right) ground truth images.}
\label{fig:clf_comp_all}
\end{figure}

As ViT provides the best performance compared to the other network backbones, we take this feature extractor and perform a thorough study by comparing MGAug with all baselines on an increasing number of augmented data. The left panel of Fig.~\ref{fig:clf_comp_all} reports our model MGAug's classification accuracy (using ViT as the network backbone) on 2D silhouette shape dataset with different number of modes, $C$, along with diverse levels of augmented images quantities (i.e., $5\times, 7\times,$ and $10\times$ the original training data size). It suggests that the utilization of a multimodal latent space $(C>1)$ consistently yields superior performance compared to an unimodal distribution $(C=1)$ by more than $2$-$7\%$. Our model achieves an optimal performance when $C=7$, which represents the number of classes in the training data. The right panel of Fig.~\ref{fig:clf_comp_all} shows a comprehensive evaluation of classification accuracy across all augmentation methods with an increased number of augmented images in training data. Our method significantly outperforms other baselines by more than $7\%$ of accuracy with different numbers of augmented images. Another interesting observation is that our MGAug achieves comparable classification accuracy when using roughly $3 \times$ augmented images for training vs. other baselines using $10\times$ of ground truth images. This indicates the potency of our method in efficiently leveraging augmented data to achieve competitive results.

To further validate the effectiveness of our model on varying sizes of ground truth, we take $75\%$ and $50\%$ of ground truth images from the 2D silhouette shape dataset and perform classification under the ViT backbone. Fig. \ref{fig:clf_comp_all} depicts a classification accuracy comparison of all the augmentation models on this setting over increasing number of augmented images. Our model still achieves best results in varying number of augmented images. This further proves the efficiency of using our model when we have incomplete/missing/limited ground truth images.

\begin{figure*}[h]
    \centering
    \includegraphics[width=0.9\linewidth]{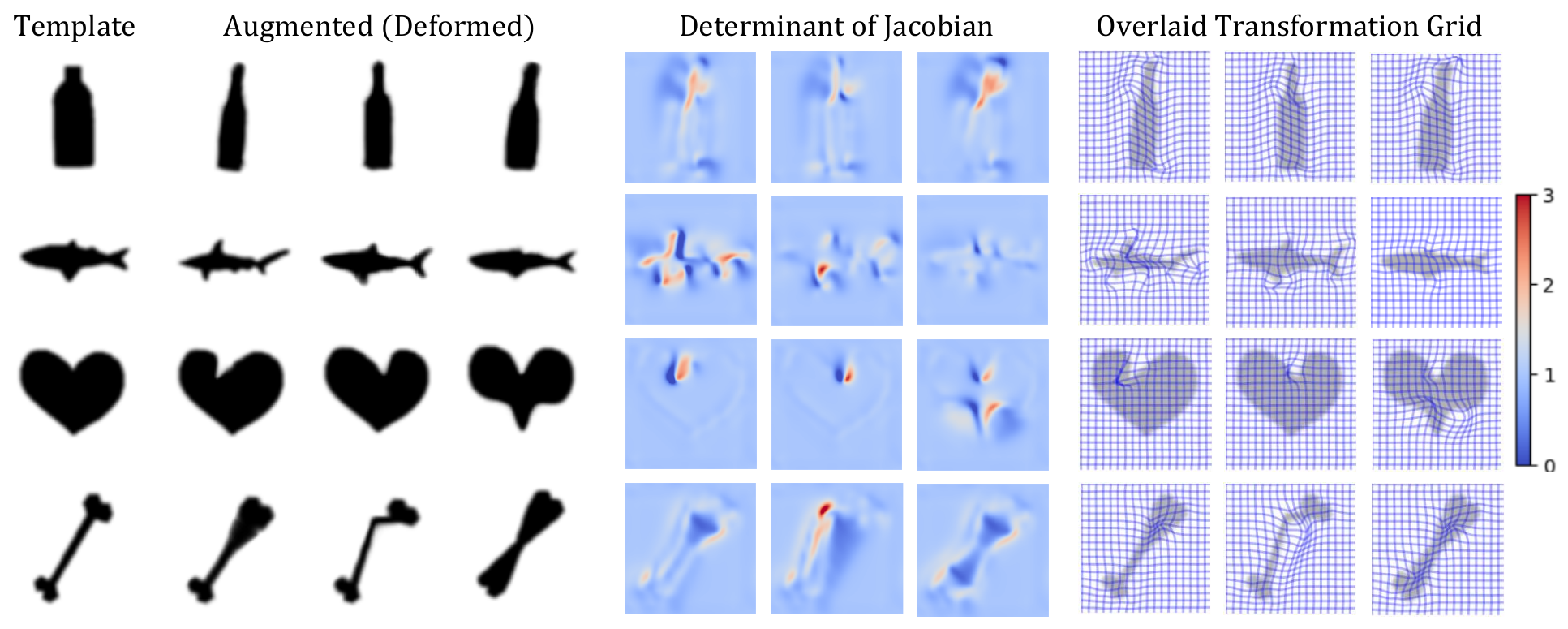} 
    \caption{Examples of DetJac generated by MGAug. Left to right: template, augmented (deformed), DetJac overlaid with augmented images.}
    \label{fig:sil_shapes}
\end{figure*}

Fig.~\ref{fig:sil_shapes} visualizes examples of augmented images and the determinant of Jacobian (DetJac) of augmented transformations produced by MGAug. The value of DetJac indicates volume changes when applying augmented deformations on templates. For example, there is no volume change when DetJac$=$1, while volume shrinks when DetJac$<$1 and expands when DetJac$>$1. The value of DetJac smaller than zero indicates an artifact or singularity in the transformation field, i.e., a failure to preserve the diffeomorphic property when the effect of folding and crossing grids occurs. Our method nicely preserves the topological structures of augmented images without introducing artifacts. 

\begin{figure}[htbp]
\centering
\begin{subfigure}{.47\linewidth}
  \centering
  \includegraphics[width=\linewidth]{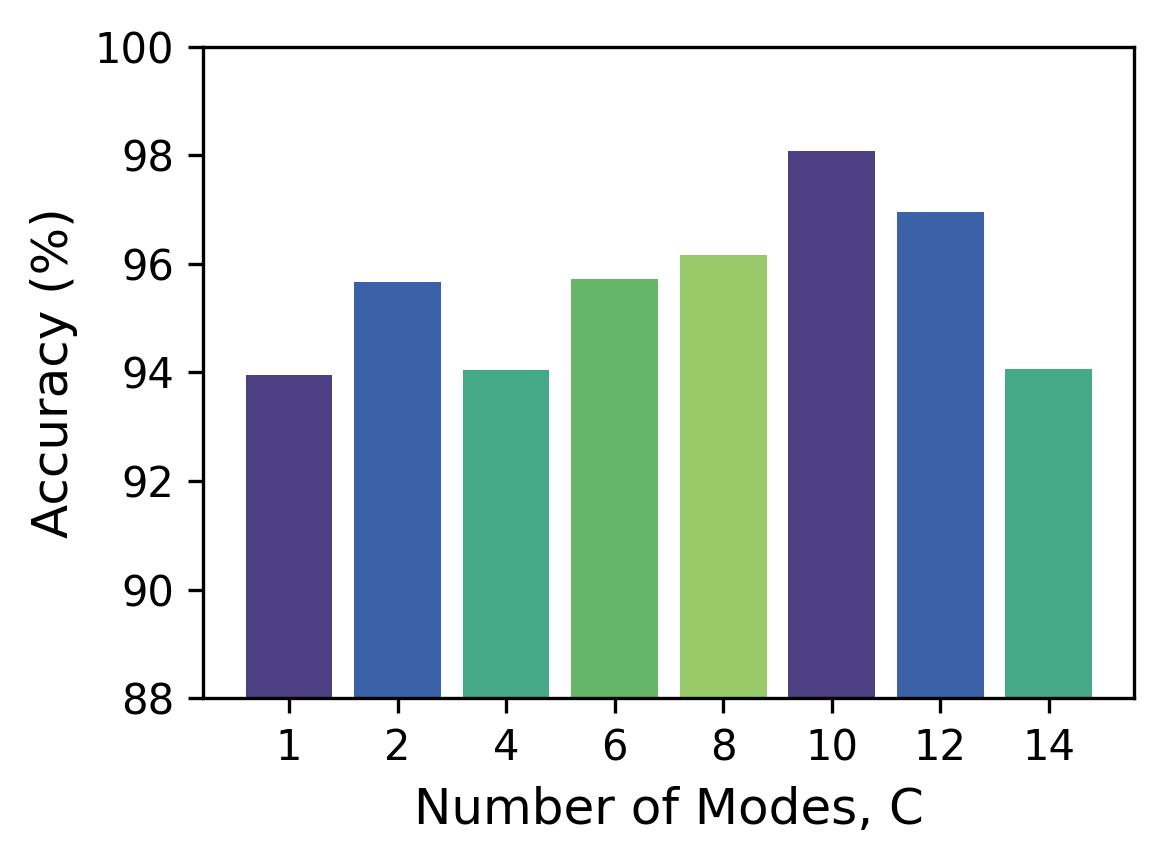}
  \caption{Cross validation on $C$}
  \label{fig:mnist-sub-first}
\end{subfigure}
\begin{subfigure}{.45\linewidth}
  \centering
  \includegraphics[width=\linewidth]{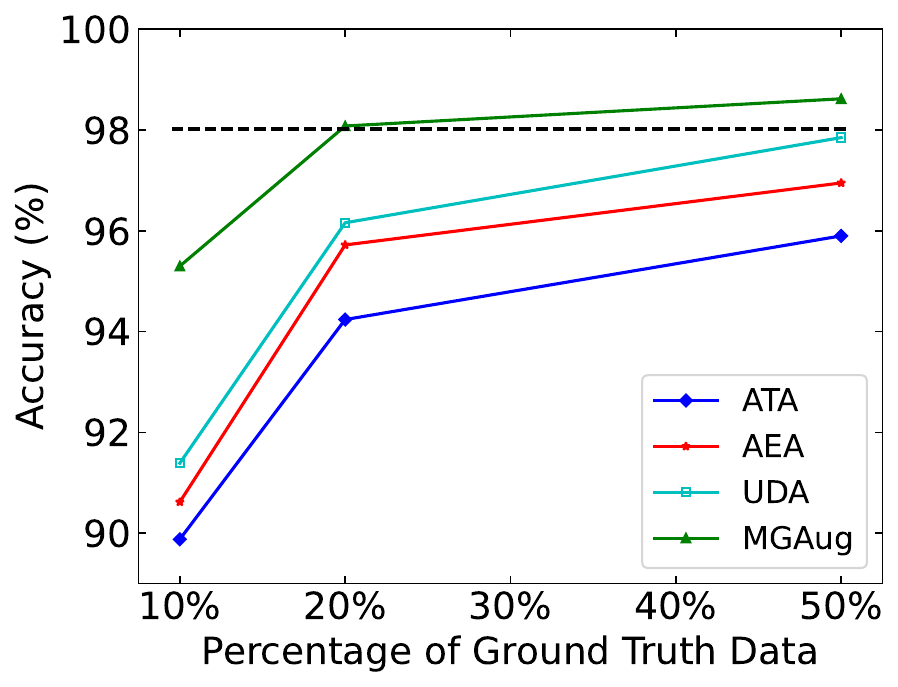} 
  \caption{Varying GT images.}
  \label{fig:mnist-sub-second}
\end{subfigure}

\vspace{0.2cm}

\begin{subfigure}{.45\linewidth}
  \centering
  \includegraphics[width=\linewidth]{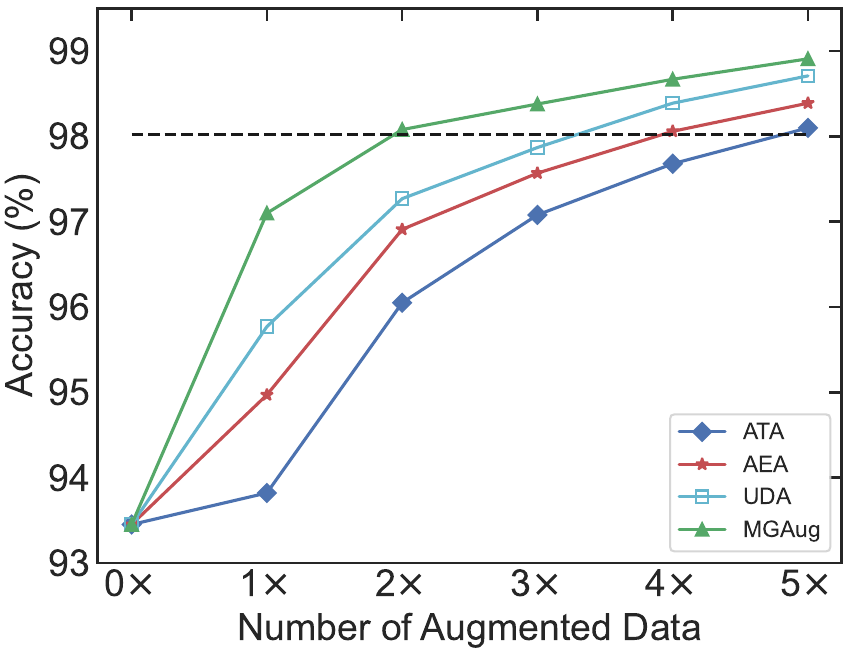} 
  \caption{Backbone: MLP}
  \label{fig:mnist-sub-third}
\end{subfigure}
\begin{subfigure}{.45\linewidth}
  \centering
  \includegraphics[width=\linewidth]{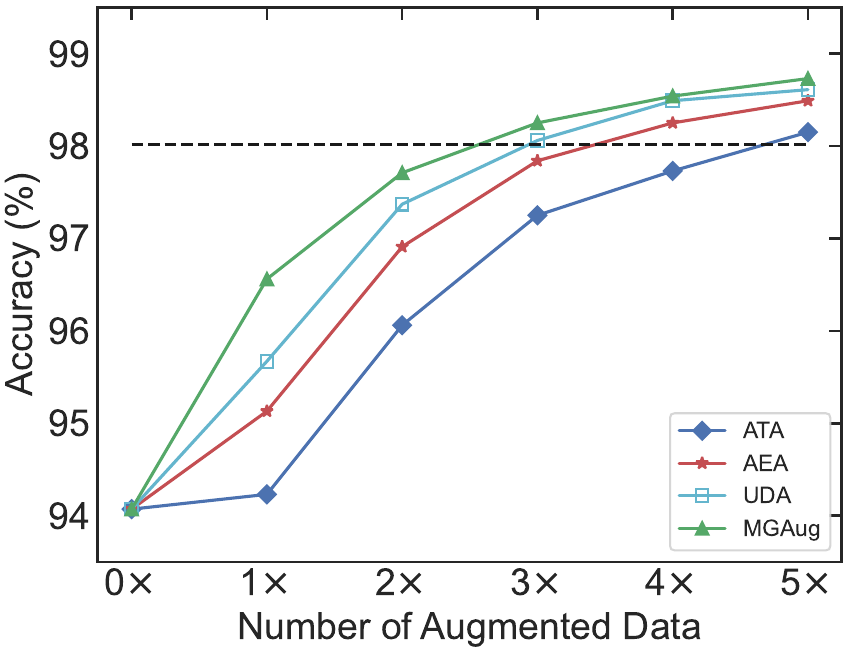} 
  \caption{Backbone: CNN}
  \label{fig:mnist-sub-fourth}
\end{subfigure}
\caption{\textbf{Top}: Accuracy comparison on various modes $C$ on 2D handwritten dataset taking $20\%$ ground-truth images over $3\times$ augmentations (left panel) and classification evaluation over increasing ground-truth images taking $3\times$ augmentations (right panel). \textbf{Bottom}: A comparison of classification performance for all models on 2D handwritten digits over an increasing amount of augmented data taking $10\%$ ground truth images under MLP (left panel) and CNN backbone (right panel). The horizontal dashed line ($\mbox{- -}$) serves as a reference, representing the classification accuracy achieved when utilizing all available training images without any augmentation.}
\label{fig:clf_comp_mnist}
\end{figure}

Fig.~\ref{fig:clf_comp_mnist} reports the ablation and classification performances on the 2D handwritten MNIST dataset. Fig.~\ref{fig:mnist-sub-first} reports the classification accuracy over an increasing number of modes, $C$. Similarly, multimodal latent space $(C>1)$ yields comparatively better performance than the unimodal distribution $C=1$ and achieves optimal accuracy when $C=10$, which represents the number of classes in the MNIST dataset. Fig.~\ref{fig:mnist-sub-second} showcases the classification results on the 2D MNIST dataset for all methods, using different percentages of ground truth data during the training phase. Our MGAug method consistently achieves the highest accuracy across all augmentation techniques. Remarkably, MGAug reaches an accuracy of $98.02\%$ (which is the baseline accuracy without any augmentation) using only $20\%$ of the ground truth data, whereas other methods require at least $50\%$ of ground truth data to achieve similar results. The bottom panel of Fig.~\ref{fig:clf_comp_mnist} reports the accuracies on the 2D MNIST dataset using $10\%$ ground truth images over an increasing number of augmented data ($1\times$ - $5\times$) under two different network backbones - MLP (Fig.~\ref{fig:mnist-sub-third}) and 2-block CNN (Fig.~\ref{fig:mnist-sub-fourth}). It shows that MGAug consistently outperforms all the baselines over any amount of augmented data settings. Moreover, MGAug reaches the classification accuracy which uses all ground truth images (without no augmentations) by using only $2\times$ and $2.5\times$ of the $10\%$ ground truth images whereas the closest unimodal model UDA requires more than $3\times$ augmented data under both MLP and CNN backbone, respectively.

Tab.~\ref{tab:mnist_tab_comp} reports the classification accuracy of unimodal UDA and multimodal MGAug under different network backbones on $2\%$ ground truth images. For this experiment, we present the results for $2\times$ and $3\times$ augmentations. We run the experiment three times and report the average accuracy and their variances. MGAug achieves the best performance under all network backbones with very little variance, whereas unimodal UDA has comparable accuracy but with higher variances. This further shows the generalizability power of our model MGAug under various feature extractor backbones.

\begin{table}[!htb]
\centering
\caption{Accuracy comparison on 2D handwritten digits over unimodal UDA model across various network backbones using $2\%$ ground truth images.}
\resizebox{\linewidth}{!}{\begin{tabular}{lcccc}
\toprule
Augmentation & \multicolumn{2}{c}{$2 \times$} & \multicolumn{2}{c}{$3 \times$} \\
\midrule
\textit{Backbones} & \textit{UDA}& \textit{MGAug}                & \textit{UDA}          & \textit{MGAug}        \\
            \midrule
ResNet18    & $97.42\pm.19$          & $97.90\pm.06$            & $97.89\pm.03$   & $\mathbf{98.10\pm.01}$   \\
ResNet34    & $97.27\pm.30$          & $93.51\pm.03$          & $94.52\pm.11$   & $\mathbf{95.48\pm.02}$   \\
VGGNet16    & $98.37\pm.19$          & $98.58\pm.08$           & $98.42\pm.06$   & $\mathbf{98.64\pm.05}$   \\
VGGNet19    & $98.29\pm.30$          & $98.38\pm.04$           & $98.57\pm.04$   & $\mathbf{98.67\pm.11}$   \\
DenseNet121 & $98.27\pm.04$          & $98.48\pm.01$           & $98.47\pm.03$   & $\mathbf{98.60\pm.02}$   \\
DenseNet169 & $98.40\pm.07$          & $98.55\pm.002$          & $98.67\pm.11$   & $\mathbf{98.78\pm.03}$ \\
\bottomrule
\end{tabular}}
\label{tab:mnist_tab_comp}
\end{table}

\begin{figure}[h]
    \centering
    \includegraphics[width=0.95\linewidth]{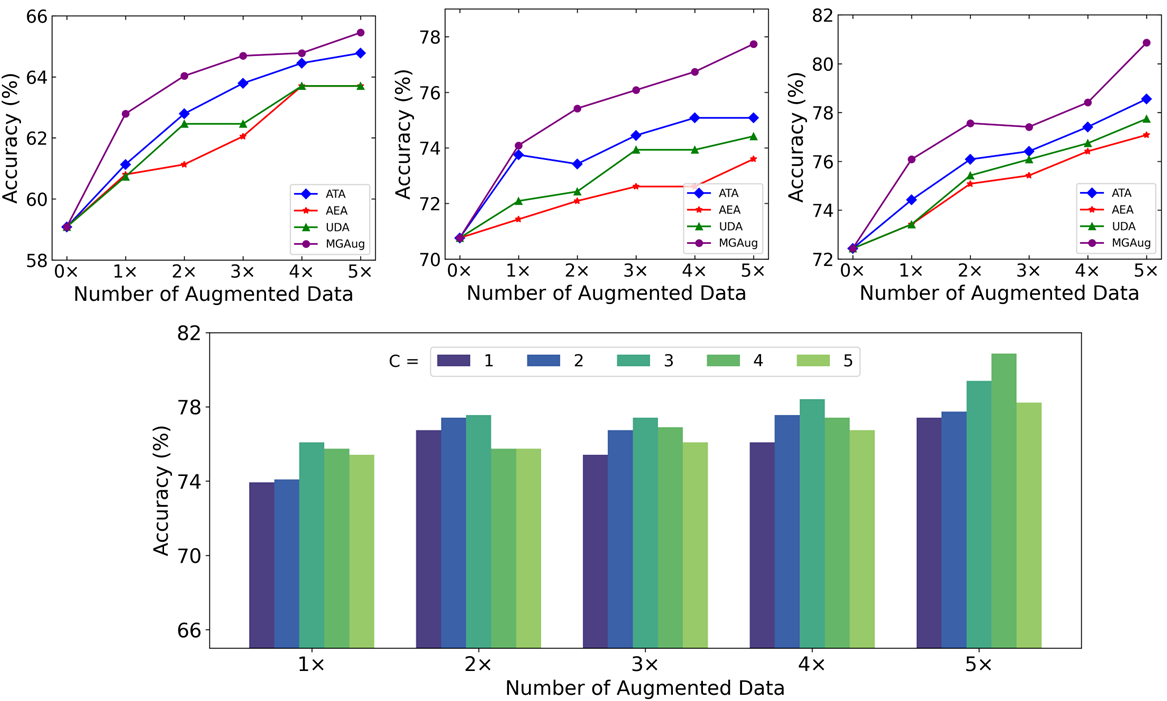} 
    \caption{Top: A comparison of classification performance for all models over increasing number of augmented data taking $25\%$ (top-left), $50\%$ (top-middle), and $75\%$ (top-right) ground truth images. Bottom: Accuracy comparison on various modes C on 3D brain MRI dataset for classification tasks.}
    \label{fig:brain_mri_clf}
\end{figure}

Fig.~\ref{fig:brain_mri_clf} (top-panel) visualizes the comparative analysis of data augmentation methods on 3D brain MRI classification. The top row presents classification accuracy trends across different proportions of ground truth data: $25\%$ (top-left), $50\%$ (top-middle), and $75\%$ (top-right) over an increased number of augmented data ($0\times$ - $5\times$) for all methods. Our model consistently demonstrates superior performance across all settings, with particularly notable improvements in scenarios with limited ground truth data ($25\%$).

Fig.~\ref{fig:brain_mri_clf} (bottom-panel) illustrates an extensive ablation study on the impact of hyperparameter $C$ (ranging from $1$ to $5$) on model accuracy using different augmentation ratios ($1\times$ to $5\times$) on the 3D brain MRI dataset taking $75\%$ ground-truth images. This ablation study reveals the sensitivity of the model to both the amount of augmented data and the choice of $C$, with optimal performance typically achieved using moderate $C$ values ($C=2$-$C=4$) and higher augmentation ratios ($3\times$-$5\times$). The results suggest that while increasing the amount of augmented data generally improves performance, the selection of an appropriate $C$ value is crucial for maximizing the benefits of the augmentation strategy in these classification tasks.

\begin{figure}[htbp]
\centering
\begin{subfigure}{.45\linewidth}
  \centering
  \includegraphics[width=\linewidth]{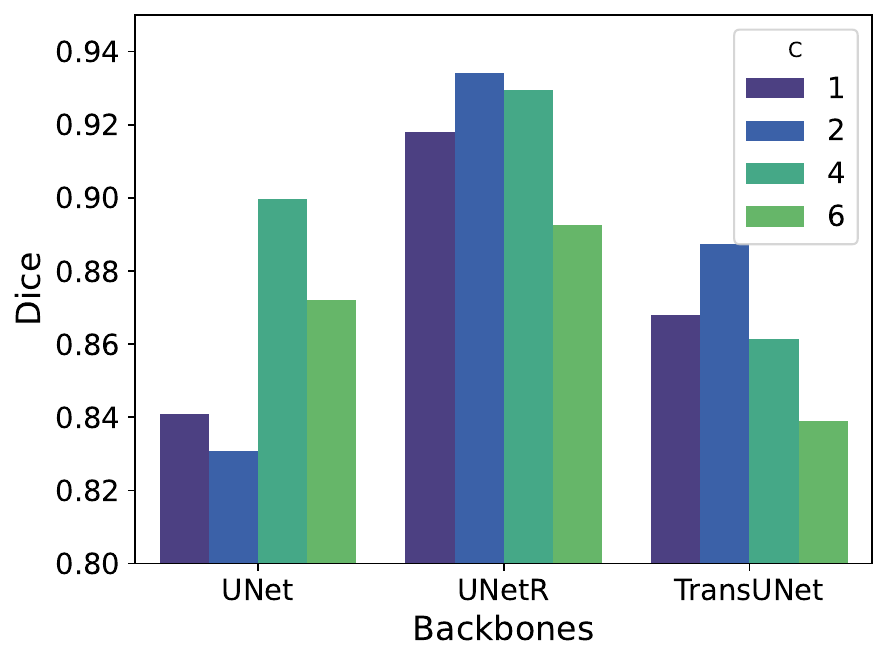} 
\end{subfigure}
\begin{subfigure}{.45\linewidth}
  \centering
  \includegraphics[width=\linewidth]{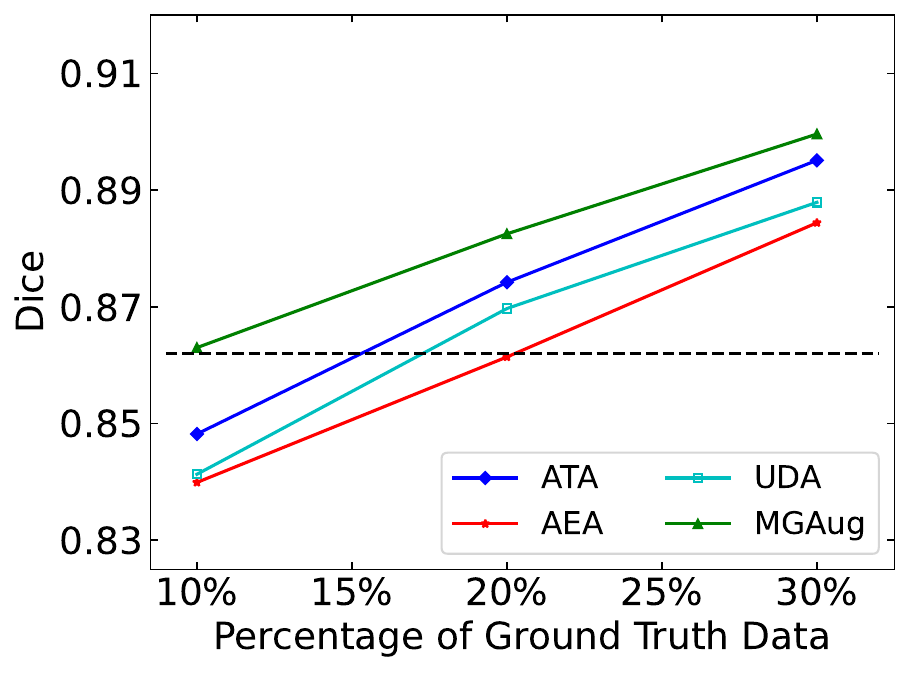} 
\end{subfigure}
\centering
\begin{subfigure}{.45\linewidth}
  \centering
  \includegraphics[width=\linewidth]{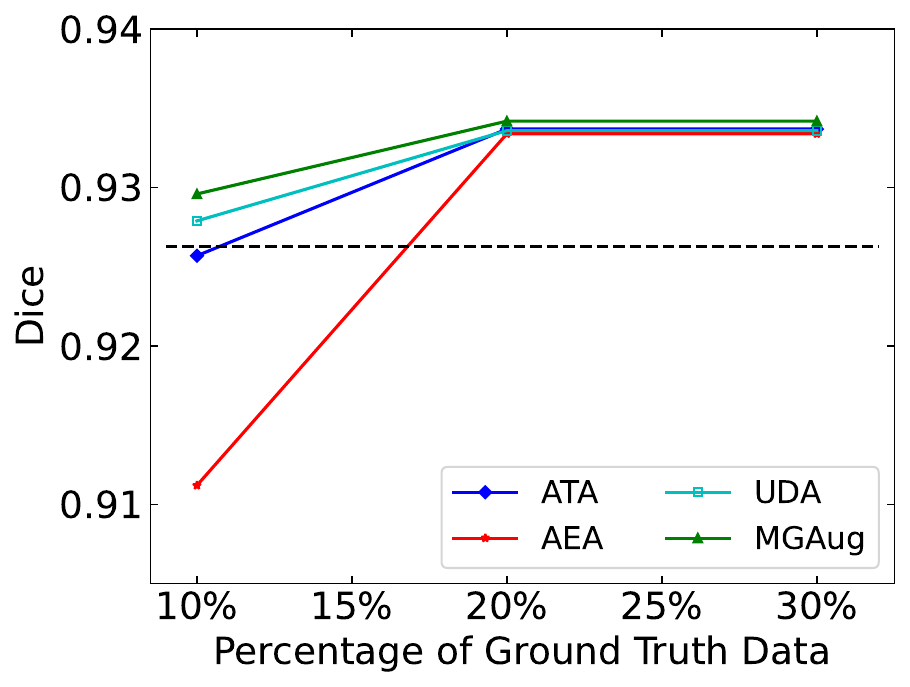} 
\end{subfigure}
\begin{subfigure}{.45\linewidth}
  \centering
  \includegraphics[width=\linewidth]{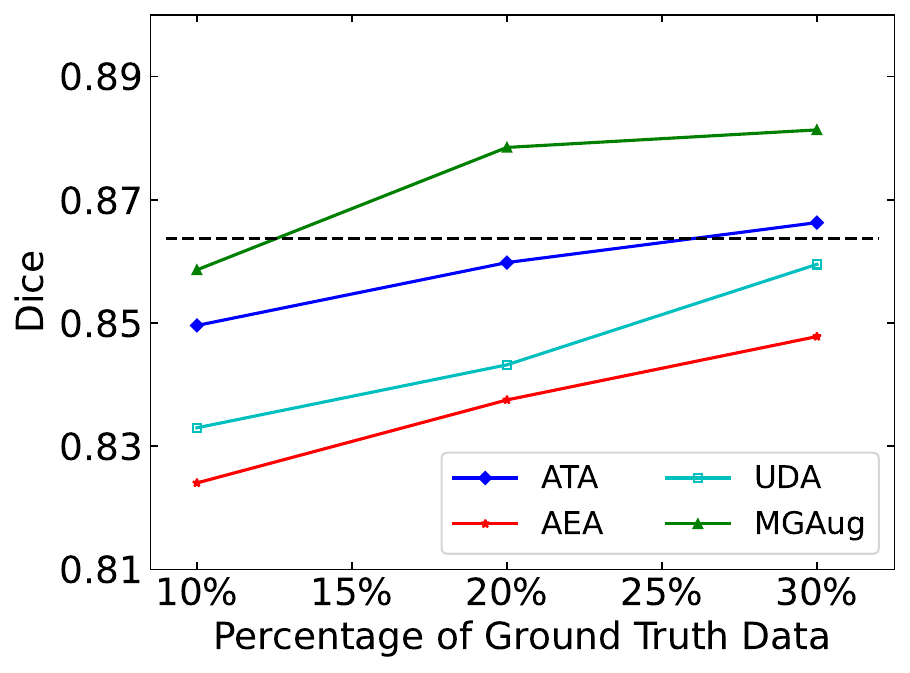} 
\end{subfigure}
\caption{Top-Left: Dice comparison on various modes $C$ on 3D brain MRI dataset. Right: A comparison of segmentation performance for all models over increasing number of ground-truth data ($10\% - 30\%$) under UNet (Top-Right), UNetR (Bottom-Left), and TransUNet (Bottom-Right) backbones. For fair comparison, we augment the data $3\times$ to the ground truth images. The horizontal dashed line ($\mbox{- -}$) serves as a reference, representing the segmentation dice score achieved when utilizing all available training images without any augmentation.}
\label{fig:seg_dice_all}
\end{figure}

Fig.~\ref{fig:seg_dice_all} presents a comprehensive evaluation of segmentation performance across different network architectures across all data augmentations on 3D brain MRI datasets along with a comprehensive ablation study on modes $C$. The top-left subplot demonstrates the ablation study of MGAug across different modes $(C=1-4)$ on three backbone architectures (UNet, UNetR, and TransUNet), where mode $C=2/4$ consistently achieves superior performance, particularly notable in UNet and UNetR backbones. The remaining three subplots showcase segmentation performance comparisons between MGAug and other baseline augmentation methods under limited ground truth scenarios $(10-30\%)$, taking $3\times$ augmentations of the ground truth data for fair comparison. The results demonstrate that MGAug achieves comparable or superior performance to this full-data baseline using only $20-25\%$ of ground truth data. For UNet (top-right panel), MGAug maintains a clear advantage over competing methods, achieving approximately $2\%$ higher Dice scores. In UNetR (bottom-left), MGAug rapidly reaches the full-data baseline performance with just $20\%$ of ground truth data. On TransUNet (bottom-right), MGAug demonstrates consistent superiority, outperforming other methods by up to $3\%$ in Dice score. This indicates MGAug's exceptional efficiency in leveraging limited ground truth data through augmentation, making it particularly valuable for real-world scenarios where labeled data is scarce.

\begin{figure*}[htbp]
    \centering
    \begin{subfigure}[b]{1.0\linewidth}
        \centering
        \includegraphics[width=\linewidth]{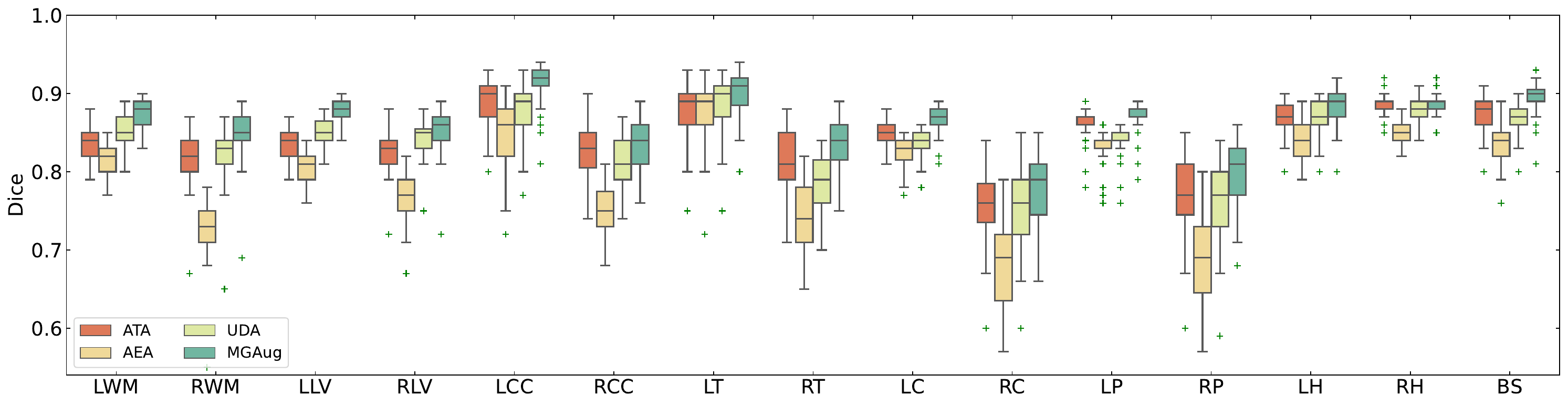}
        \label{fig:seg_ex_dice_1}
    \end{subfigure}
    \begin{subfigure}[b]{1.0\linewidth}
        \centering
        \includegraphics[width=\linewidth]{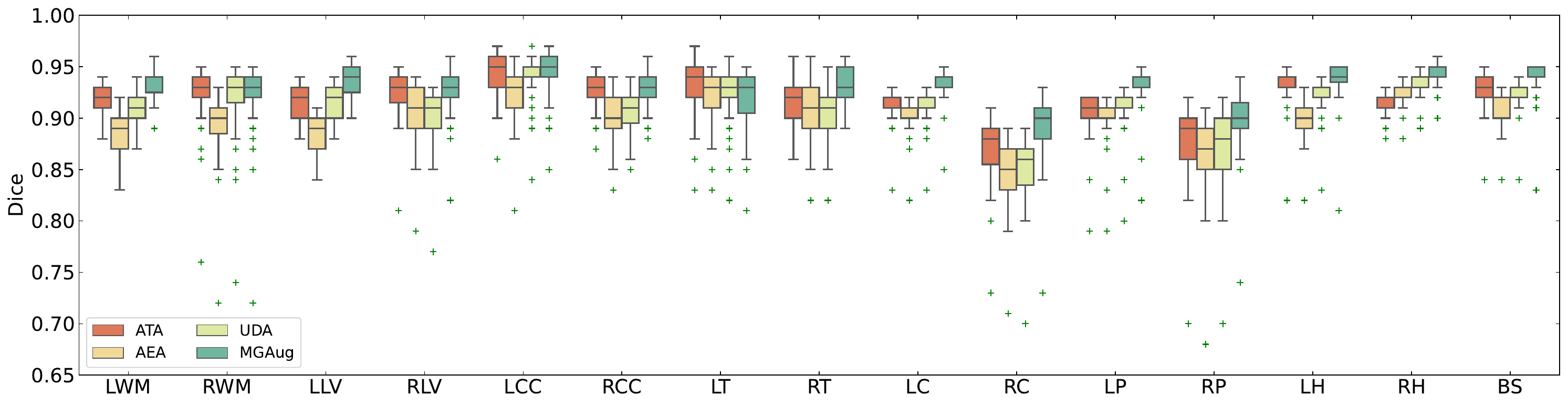}
        \label{fig:seg_ex_dice_2}
    \end{subfigure}
    \begin{subfigure}[b]{1.0\linewidth}
        \centering
        \includegraphics[width=\linewidth]{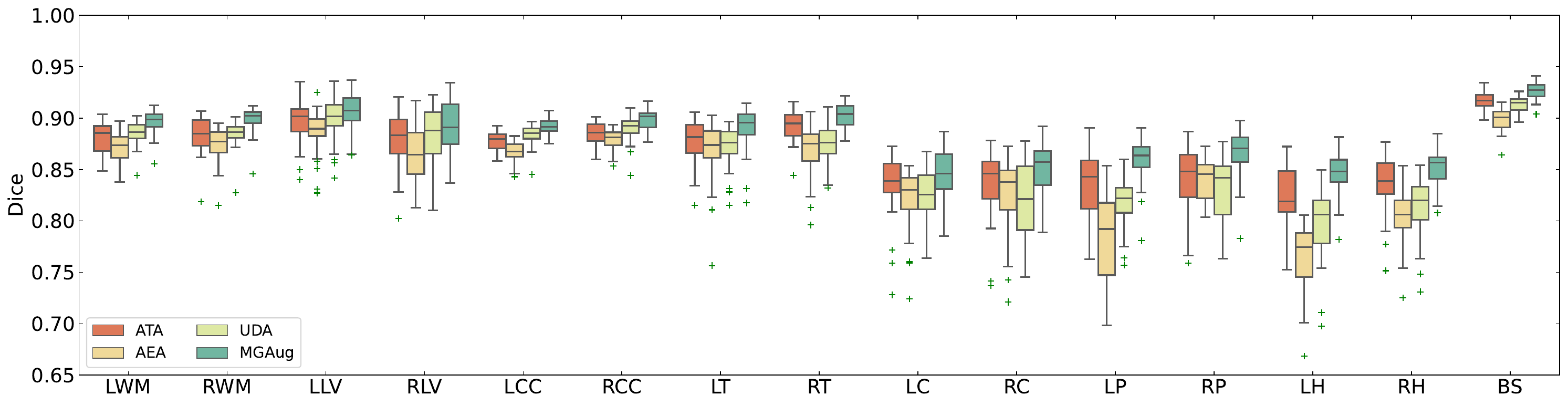}
        \label{fig:seg_ex_dice_3}
    \end{subfigure}
    \caption{Dice score comparison over fifteen anatomical brain structures (LWM - Left White Matter, RWM - Right White Matter, LLV - Left Lateral Ventricle, RLV - Right Lateral Ventricle, LCC - Left Cerebellum Cortex, RCC - Right Cerebellum Cortex, LT - Left Thalamus, RT - Right Thalamus, LC - Left Cudate, RC - Right Caudate, LP - Left Putamen, RP - Right Putamen, LH - Left Hippocampus, RH - Right Hippocampus, BS - Brain Stem) for four augmentation methods on $3\times$ augmentations under UNet (top), UNetR (middle), and TransUNet (bottom) backbone. For fair comparison, we use $10\%/20\%/30\%$ ground-truth images for UNet/UNetR/TransUNet backbones, respectively.}
    \label{fig:seg_ex_dice}
\end{figure*}

Fig.~\ref{fig:seg_ex_dice} reports the statistics of Dice scores (mean and variance) of all augmentation methods over fifteen anatomical brain structures using UNet (top), UNetR (middle), and TransUNet (bottom) backbones. For fair comparison, we use $10\%/20\%/30\%$ ground-truth images for UNet/UNetR/TransUNet backbones, respectively. Our MGAug consistently outperforms baseline methods (ATA, UDA, AEA) across most anatomical structures, particularly showing notable improvements in challenging regions like Left/Right Caudate (LC/RC) and Left/Right Putamen (LP/RP). The box plots further demonstrate MGAug's superior performance with generally higher mean segmentation dice and comparable or lower variance across structures, though some outliers are observed in certain regions.

\begin{figure}[htbp]
    \centering
    \includegraphics[width=\linewidth]{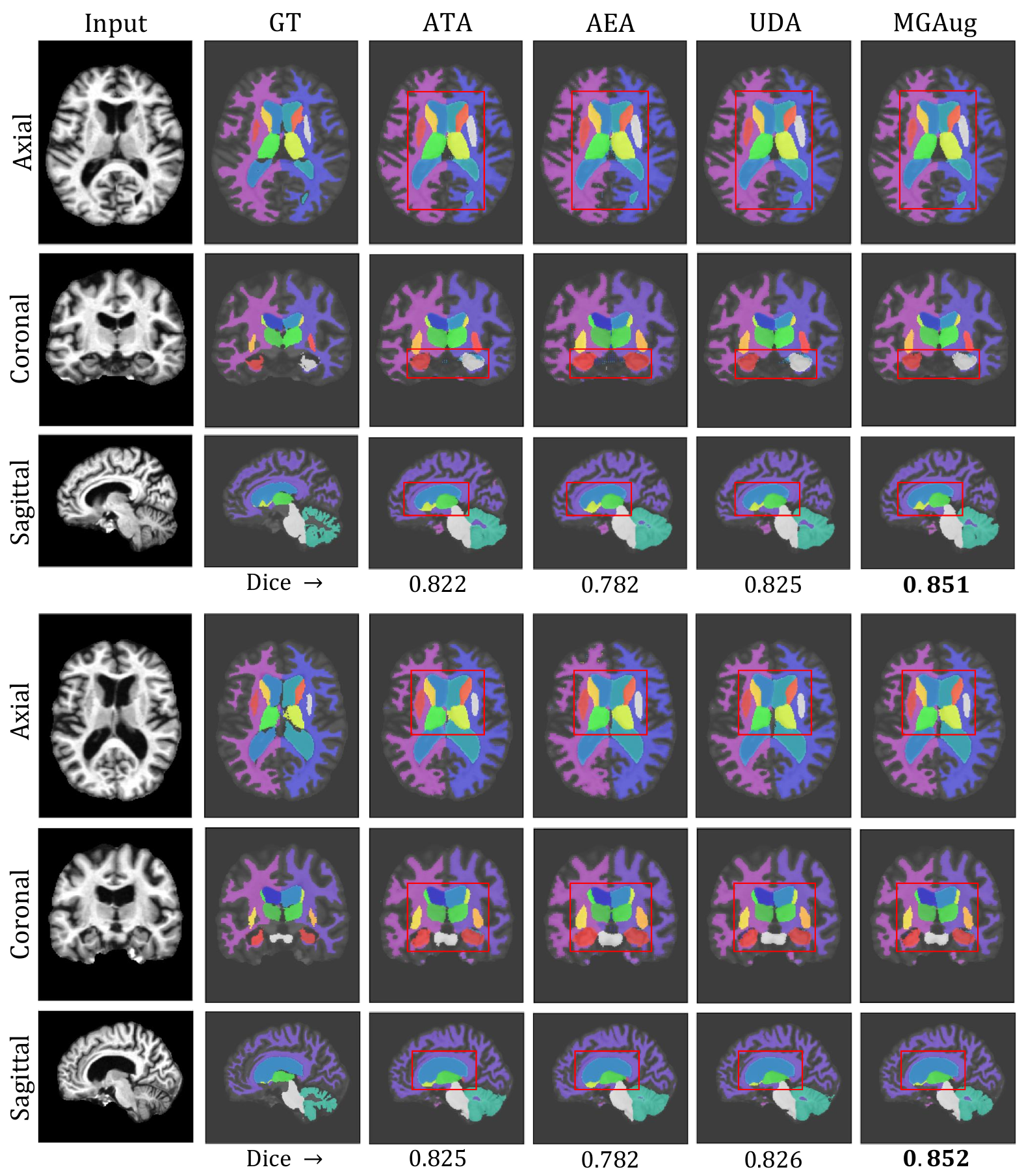}
    \caption{Left to Right: Examples of segmentation (overlaid with input images) comparisons between the ground truth (GT) and predictions of all baselines including MGAug on 3D brain MRIs over three anatomical views (Top to Bottom: Axial, Coronal, and Sagittal).}
    \label{fig:brain_seg}
\end{figure}

Fig.~\ref{fig:brain_seg} presents brain MRI segmentation results across axial, coronal, and sagittal views, comparing manually delineated ground truth with predictions from all models. It shows that MGAug achieves  superior performance than baselines.

\begin{figure}[htbp]
    \centering
    \includegraphics[width=\linewidth]{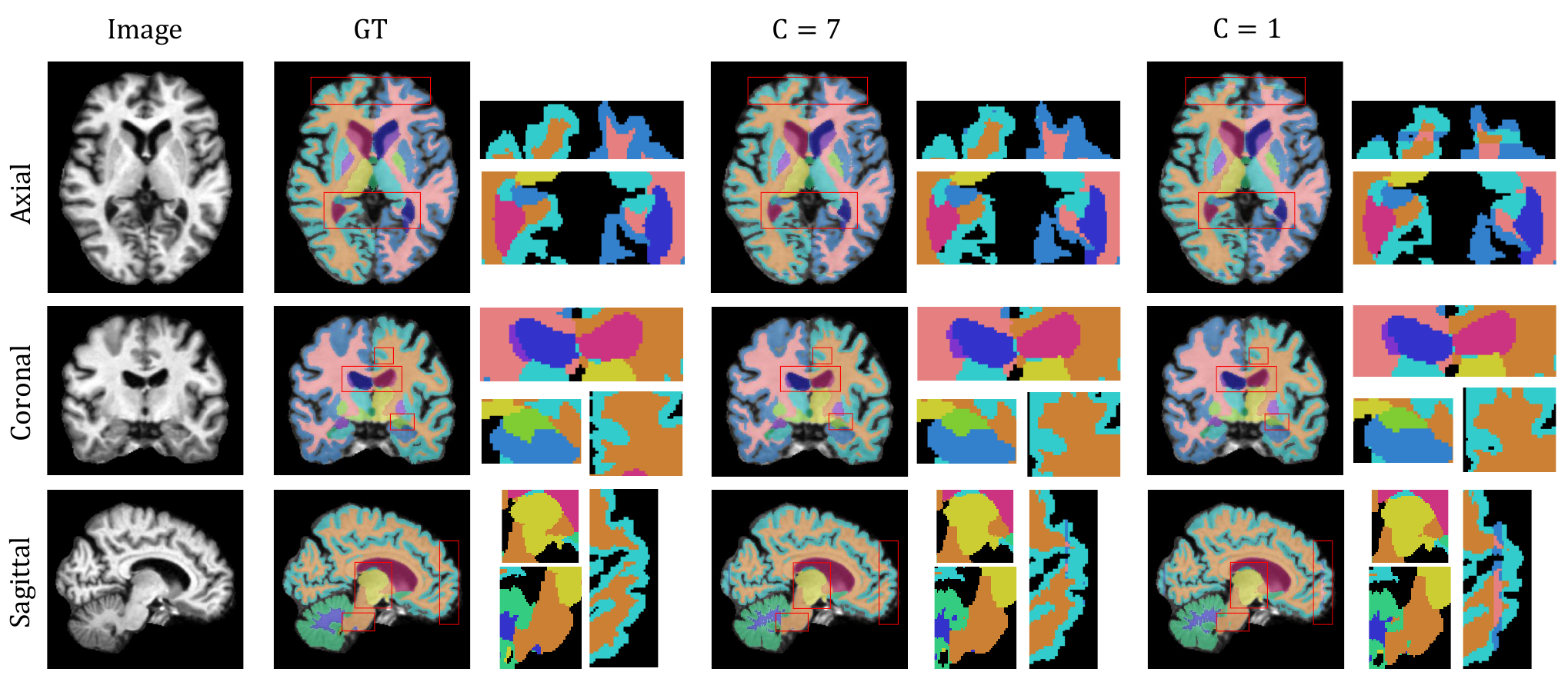}
    \caption{Left to Right: Segmentation comparisons between ground truth (GT), multimodal ($C=7$) and unimodal ($C=1$) MGAug with a zoomed-in view, over three anatomical views (Top to Bottom: Axial, Coronal, and Sagittal).}
    \label{fig:brain_seg_modes}
\end{figure}

Fig.~\ref{fig:brain_seg_modes} visualizes segmentation comparisons of our model ($C=7$) vs. unimodal ($C=1$) across three anatomical views. We highlight small anatomical structures (outlined in redbox) that show improved performance, demonstrating the superior segmentation quality of our approach.

\paragraph*{\bf MGAug vs. random/adversarial deformable augmentation}

Tab.~\ref{tab:def_combined} presents performance evaluations comparing three MGAug with random and adversarial augmentation models on multi-class classification and segmentation tasks using 3D real brain MRI dataset. In the classification task (Tab.~\ref{tab:clf_def}), we conduct experiments with varying ground-truth (GT) percentages and augmentation ratios under the ResNet backbone. MGAug consistently outperforms both Random and AdvChain approaches across all settings, demonstrating superior performance with higher GT percentages and increased augmentation ratios. Similarly, in the segmentation task (Tab.~\ref{tab:seg_def}) under UNet backbone with varying GT images with $3\times$ augmentations, MGAug demonstrates consistently better performance across different GT, achieving higher dice scores compared to the baseline methods. These results indicate MGAug's effectiveness in both classification and segmentation tasks, particularly in scenarios with limited ground-truth data.

\begin{table}[htbp]
\centering
\caption{Performance comparison of random deformable augmentation (Random), adversarial deformable augmentation (AdvChain), and MGAug on (a) multi-class classification and (b) segmentation dice on 3D brain MRI dataset.}
\captionsetup[subtable]{font=small}  
\label{tab:def_combined}

\begin{subtable}{\linewidth}
\centering
\caption{Classification accuracy comparison for $50\%$ and $75\%$ ground-truth (GT) images with an increasing number of augmentations ($1\times$ to $5\times$) under the ResNet backbone. GT: Ground-truth images.}
\begin{tabular}{lcccc}
\toprule
\multicolumn{1}{l}{GT} & \multicolumn{1}{l}{Augmentations} & Random & AdvChain & MGAug \\ \midrule
\multirow{5}{*}{$50\%$} & $1\times$                                & $71.76$  & $73.25$    & $\mathbf{74.09}$ \\ \cmidrule{2-5}
                        & $2\times$                                & $72.76$  & $74.75$    & $\mathbf{75.42}$ \\ \cmidrule{2-5}
                        & $3\times$                                & $73.10$   & $75.08$    & $\mathbf{76.08}$ \\ \cmidrule{2-5}
                        & $4\times$                                & $74.09$  & $75.25$    & $\mathbf{76.74}$ \\ \cmidrule{2-5}
                        & $5\times$                                & $75.08$  & $76.24$    & $\mathbf{77.74}$ \\ \midrule
\multirow{5}{*}{$75\%$} & $1\times$                                & $74.42$  & $75.75$    & $\mathbf{76.08}$ \\ \cmidrule{2-5}
                        & $2\times$                                & $75.08$  & $76.08$    & $\mathbf{77.56}$ \\ \cmidrule{2-5}
                        & $3\times$                                & $76.24$  & $77.08$    & $\mathbf{77.41}$ \\ \cmidrule{2-5}
                        & $4\times$                                & $76.41$  & $77.23$    & $\mathbf{78.41}$ \\ \cmidrule{2-5}
                        & $5\times$                                & $77.74$  & $79.07$    & $\mathbf{80.86}$ \\ \bottomrule
\end{tabular}
\label{tab:clf_def}
\end{subtable}

\bigskip

\begin{subtable}{\linewidth}
\centering
\caption{Segmentation dice score comparison for increasing ground-truth (GT) images ($10\%$ to $30\%$) with $3\times$ augmentations under the UNet backbone. GT: Ground-truth images}
\begin{tabular}{lccc}
\toprule
\multicolumn{1}{l}{GT} & Random & AdvChain & MGAug \\ \midrule
$10\%$ & $82.65$  & $84.91$    & $\mathbf{86.30}$  \\ \midrule
$20\%$ & $84.13$  & $86.97$    & $\mathbf{88.25}$ \\ \midrule
$30\%$ & $88.25$  & $88.44$    & $\mathbf{89.96}$ \\ \bottomrule
\end{tabular}
\label{tab:seg_def}
\end{subtable}
\end{table}

\paragraph*{\bf MGAug vs. intensity-based augmentations} While the primary emphasis of this paper does not fall into the realm of intensity-based augmentations, we compare MGAug with a few notable DA schemes manipulating image intensities or textures. These methods include \textit{AutoAugment}, an automated scheme that searches for improved augmentation policies~\cite{cubuk2019autoaugment}; \textit{RandAugment}, an augmentation algorithm which reduces the computational expense by a simplified search space~\cite{cubuk2020randaugment}; \textit{AugMix}, a data processing method to improve the robustness and uncertainty estimates~\cite{hendrycks2019augmix}; and \textit{TrivialAugment}, a parameter-free data augmentation model~\cite{muller2021trivialaugment}. We maintain the same training protocols as the original models. For a fair comparison, we augment the data $10\times$ times for 2D shapes (with all training samples) and $3\times$ times for 2D digits (with $50\%$ of the training images) experiments.

Tab.~\ref{comp_intensity} reports the accuracy comparison of a variety of DA schemes for image classification networks based on the ViT backbone. The testing results on both MNIST digits and 2D shape data show that our model outperforms the baselines consistently.
\begin{table}[!h]
\caption{Accuracy comparison on both 2D digits and 2D shape datasets over a mix of augmentation methods (including intensity-based vs. transformation-based DA) on ViT backbone.}
  \centering
\begin{tabular}{lcccc}
\toprule
                 \textbf{Datasets}              & \multicolumn{2}{c}{\textbf{2D Digits}} & \multicolumn{2}{c}{\textbf{2D Shape}} \\
                  \cmidrule{1-5}
                  Models         & \textit{Acc}          & \textit{F1}     & \textit{Acc}         & \textit{F1}     \\
                  \midrule
AutoAugment    & 90.82                & \multicolumn{1}{c|}{90.73}          &  76.05        & 75.25       \\
                  RandAugment    &  95.25               &  \multicolumn{1}{c|}{95.22}        &  81.25        & 81.70       \\
                  AugMix                  &   97.31     & \multicolumn{1}{c|}{97.30}          & 74.28         & 69.35        \\
                  TrivialAug &  95.48                &  \multicolumn{1}{c|}{95.49}        &  83.93        &  82.75      \\
                  \midrule
Affine         & 95.90                  &  \multicolumn{1}{c|}{95.84}      & 78.57                & 76.53       \\
                  AEA      &  96.95                 &  \multicolumn{1}{c|}{96.91}      & 89.28                   & 89.33       \\
                  UDA            &  \underline{97.85}      &  \multicolumn{1}{c|}{\underline{97.83}}      &  \underline{91.07}               & \underline{96.84}        \\
                  MGAug          &  \textbf{98.62}     &  \multicolumn{1}{c|}{\textbf{98.78}}      & \textbf{98.21}                 & \textbf{98.20}       \\
                  \bottomrule
\end{tabular}
\label{comp_intensity}
\end{table}

\paragraph*{\bf Statistical tests}

Tab.~\ref{tab:stat} presents the statistical significance of MGAug's performance through p-values from paired t-tests against three baseline methods (ATA, AEA, UDA). The left subtable shows that MGAug achieves statistically significant improvements (p $< 0.05$) across all three segmentation backbones, with particularly strong significance for Unet (p $< 0.003$) and UNetR (p $< 0.04$). Similarly, the right subtable demonstrates that MGAug maintains statistical significance across different ground truth percentages ($25\%$, $50\%$, $75\%$) for classification tasks, with p-values consistently below $0.05$.

\begin{table}[h]
\centering
\begin{tabular}{cccc}
\hline
Model & ATA &  AEA & UDA \\
\hline
Unet & $0.0025$ & $0.0023$ & $0.0123$ \\
UNetR & $0.0076$ & $0.0034$ & $0.0370$ \\
TransUNet & $0.0343$ & $0.0197$ & $0.0537$ \\
\hline
\end{tabular}

\vspace{0.5cm} 

\begin{tabular}{cccc}
\hline
GT ($\%$) & ATA & AEA & UDA \\
\hline
$25$ & $0.037$ & $0.003$ & $0.048$ \\
$50$ & $0.027$ & $0.012$ & $0.044$ \\
$75$ & $0.048$ & $0.030$ & $0.018$ \\
\hline
\end{tabular}
\caption{Statistical test of MGAug: (Top) P-values from paired t-tests comparing MGAug with baseline methods performing segmentation on 3D brain MRIs under different segmentation backbones; (Bottom) P-values for different ground truth percentages performing classification on 3D brain MRIs.}
\label{tab:stat}
\end{table}

\section{Conclusion}
\label{sec:con}
In this paper, we present a novel data augmentation model, MGAug, which learns augmenting transformations in a multimodal latent space of geometric deformations. In contrast to existing augmentation models that adopt a unimodal distribution to augment data, our augmentation method fully characterizes the multimodal distribution of image transformations via introducing a mixture of multivariate Gaussians as the prior. Our method significantly improves the performance of multi-class classification (on 2D datasets) and image segmentation (on 3D brain MRIs) and eliminates the artifacts of geometric transformations by preserving the topological structures of augmented images. While MGAug focuses on LDDMM, the theoretical tool developed in our model is widely applicable to various registration frameworks, for example, stationary velocity fields~\cite{modat2012parametric,hadj2016longitudinal}.

To the best of our knowledge, we are the first to learn geometric deformations in a multimodal manner via deep neural networks. Our augmentation model can be easily applied to different domain-specific tasks, e.g., image reconstruction and object tracking. Interesting future directions based on this work can be i) investigating the model interpretability for classification tasks when geometric deformations are learned in a multimodal manner, and ii) the robustness study of MGAug when image intensity variations occur, i.e., under adversarial noise attacks. 

\paragraph*{\bf Acknowledgment} This work was supported by NSF CAREER Grant 2239977.

\bibliographystyle{abbrv}
\bibliography{references}

\end{document}